\documentclass[11pt]{article}
\usepackage{amsfonts, amssymb, amsmath, amsthm, enumerate, xspace, array, multirow, graphicx, bbm, mathrsfs, tikz-cd, wasysym, mathtools, esint}
\usepackage{xcolor}
\usepackage{hyperref}
\usepackage[ruled,linesnumbered]{algorithm2e}
\hypersetup{
    colorlinks=true,
    linkcolor=blue,
    filecolor=blue,      
    urlcolor=blue,
    allcolors=blue
}
\numberwithin{equation}{section}
\usepackage[left=1 in, right=1 in, top=1in, bottom=1in]{geometry}
\usepackage{macros}
\usepackage[font=small,labelfont=bf]{caption}
\usepackage{tabularx}
\newcolumntype{C}{ >{\centering\arraybackslash}X}
\newcolumntype{B}{ >{\arraybackslash}X}
\begin{document}

\title{Random Feature Models for Learning \\Interacting Dynamical Systems}
\author{Yuxuan Liu, Scott G. McCalla, Hayden Schaeffer}\date{}

\maketitle

\begin{abstract}
Particle dynamics and multi-agent systems provide accurate dynamical models for studying and forecasting the behavior of complex interacting systems. They often take the form of a high-dimensional system of differential equations parameterized by an interaction kernel that models the underlying attractive or repulsive forces between agents. We consider the problem of constructing a data-based approximation of the interacting forces directly from noisy observations of the paths of the agents in time. The learned interaction kernels are then used to predict the agents behavior over a longer time interval. The approximation developed in this work uses a randomized feature algorithm and a sparse randomized feature approach. Sparsity-promoting regression provides a mechanism for pruning the randomly generated features which was observed to be beneficial when one has limited data, in particular, leading to less overfitting than other approaches. In addition, imposing sparsity reduces the kernel evaluation cost which significantly lowers the simulation cost for forecasting the multi-agent systems. Our method is applied to various examples, including first-order systems with  homogeneous and heterogeneous interactions, second order homogeneous systems, and a new sheep swarming system.
\end{abstract}

\section{Introduction} 
Agent-based dynamics typically employ systems of ordinary differential equations (ODEs) to model a wide range of complex behaviors such as particle dynamics, multi-body celestial mechanics, flocking dynamics, species interactions, opinion dynamics, and many other physical or biological systems. Fundamentally, agent-based dynamical systems model the collective behavior between multiple objects of interest by using a kernel that represent their interactions and thus provide equations that govern their motion.  They are able to reproduce a variety of collective phenomena with a minimal set of parameters which make them especially accessible for modelers \cite{chen2014minimal,motsch2011new,von2016anisotropic,mackey2014two,evers2015anisotropic}.  Such models of collective behavior are numerically slow to solve and difficult to analyze which often leads to issues in trying to extract parameters, i.e. the kernels.  This difficulty is typically circumvented by passing to continuum limits in order to aid computations \cite{wang2017efficient} and understand facets of the dynamics of the discrete system such as emergent patterning \cite{kolokolnikov2011stability,bertozzi2015ring,von2012predicting} or properties of the asymptotic dynamics \cite{carrillo2010asymptotic, barbaro2016phase,bertozzi2016regularity,morales2019flocking,tadmor2014critical,ha2008particle,balague2013dimensionality}. These interaction kernels provide information on the behavior between agents as well as the agents' self-driven forces. However, in practice these kernels may not be fully known and thus one needs to be able to approximate them from observation of the agents' states. In this work, we develop a sparse random feature model that uses a randomized Gaussian basis to approximate the interaction kernels from the data and use it to forecast future states and behavior. 

Learning dynamical systems from data is an important modeling problem in which one approximates the underlying equations of motion governing the evolution of some unknown system \cite{bongard2007automated, schmidt2009distilling}. This is essentially a model selection and parameter estimation problem, where the equations of interest are those that define a differential equation. One of the main challenges is to construct methods that do not overfit on the data, can handle noise and outliers, and can approximate a wide enough class of dynamics. While there are many works in the last decade on this topic, we highlight some of the related work along the direction of interpretable models, i.e. those that provide not only an approximation but a meaningful system of equations.  For a detailed overview of the approximation and inference of physical equations from data see \cite{ghattas2021learning} and the citations within. The SINDy algorithm \cite{brunton2016discovering} is a popular method for extracting governing equations from data using a sparsity-promoting method. The key idea is learn a sparse representation for the dynamical system using a dictionary of candidate functions (e.g. polynomials or trigonometric functions) applied to the data. This approach has lead to many related algorithms and applications, see \cite{champion2019data, zhang2019convergence,rudy2019data, kaiser2018sparse, hoffmann2019reactive, shea2021sindy, messenger2021weak, messenger2022online} and the reference within. The sparse optimization framework for learning governing equations was proposed in \cite{schaeffer2017learning} along with an approach to discovering partial differential equations using a dictionary of derivatives. The sparse optimization and compressive sensing approach for learning governing equations include $\ell^1$ based approaches for high-dimensional ODE \cite{schaeffer2020extracting, schaeffer2018extracting, schaeffer2017learning2} and noisy robust $\ell^0$ approaches using the weak form \cite{schaeffer2017sparse}. Another approach is the operator inference technique \cite{peherstorfer2016data} which can be used to approximate high-dimensional dynamics by learning the ODEs that govern the coefficients of the dynamics projected onto a linear subspace of small dimension.  This results in a model for the governing equation in the reduced space that does not require computing terms in the full dimension. While the methods of \cite{brunton2016discovering, schaeffer2017learning, peherstorfer2016data, mangan2019model} can be applied to a wide range of problems, they are not optimal for the problems considered here.

A nonparametric approach for learning multi-agent systems was developed in \cite{lu2019nonparametric} based on the assumption that the interaction kernel $g:\mathbb{R}_+\rightarrow \mathbb{R}$ is a function of the pairwise distances between agents. Their method builds a piecewise polynomial approximation of the interaction kernel by partitioning the set of possible distances (i.e. the input space for $g$) and (locally) estimating the kernel using the least squares method (see Section \ref{sec:model}). The training problem is similar in setup to other learning approaches for dynamical systems  \cite{peherstorfer2016data, brunton2016discovering, tran2017exact, schaeffer2017learning} which use trajectory based regression; however, since the ODE takes a special form, the training problem becomes a one dimensional regression problem and thus piecewise estimators do not incur the curse-of-dimensionality.  The theoretical results in \cite{lu2019nonparametric} focused on the first-order homogeneous case (one type of agent), but recent works generalize and extend the theory and numerical model to heterogeneous systems (i.e. different types of agents and different interaction kernels) \cite{lu2021learning1}, stochastic systems \cite{lu2021learning2}, and second-order systems \cite{miller2020learning}. In \cite{zhong2020data}, the accuracy and consistency of the approach from \cite{lu2019nonparametric} was numerically verified on a range of synthetic examples, including predicting emergent behaviors at large timescales.  A theoretical analysis of the identifiability of interaction systems was presented in \cite{li2021identifiability} which relies on a coercivity condition that is related to the conditions in  \cite{lu2019nonparametric, lu2021learning1, lu2021learning2}.

In this work, we will construct a random feature model (RFM) for learning the interaction kernel and thus the governing system for multi-agent dynamics. RFMs are a class of  nonparametric methods used in machine learning  \cite{rahimi2007random, rahimi2008uniform, rahimi2008weighted} which are related to kernel approximations and shallow neural networks.  The standard RFM is a shallow two-layer fully connected neural network whose (sole) hidden layer is randomized and then fixed \cite{rahimi2007random, rahimi2008uniform, rahimi2008weighted} while its output layer is trained with some optimization routine. From the perspective of dictionary learning, a RFM uses a set of candidate functions that are parameterized by a random weight vector as opposed to the standard polynomial or trigonometric basis whose constructions use a fixed set of functions. Theoretical results on RFM establish various error estimates, including, uniform error estimates \cite{rahimi2008uniform}, generalization bounds related to an $L^\infty$ like space \cite{rahimi2008weighted}, and generalization bounds for reproducing kernel Hilbert spaces \cite{rudi2017generalization, li2019towards, mei2021generalization, bach2017equivalence}. A detailed comparison between the approaches, training problems, and results can be found in \cite{liu2020random}.

\section{Learning Multi-Agent Systems} \label{sec:model}

Consider the $n$-agent interacting system, define the $n$ time-dependent state variables $\{\mathbf{x}_i(t)\}_{i=1}^n$,  $\mathbf{x}_i(t) \in \mathbb{R}^d$ for all $1\leq i \leq n$, whose dynamics are governed by the following system of ordinary differential equations (ODEs): 
\begin{align} \label{evolution_ODE}
\begin{cases}
    &\frac{d}{dt} {\mathbf x}_i(t) = \frac{1}{n} \sum\limits_{i'=1}^n g(\Vert    \mathbf{r}_{i',i}(t)\Vert_2)\, \mathbf{r}_{i',i}(t)\\
    &\mathbf{r}_{i',i}(t) =  \mathbf{x}_{i'}(t)-\mathbf{x}_i(t)
    \end{cases}
\end{align}
where the interaction kernel is $g:\mathbb{R}_+\to \mathbb{R}$. The velocity of each agent $\mathbf{x}_i(t)$ defined by \eqref{evolution_ODE} depends on the $g$-weighted average of distances to all other agents. Note that the velocity only depends on the relative distances between agents and not on their overall locations in space or time. We will assume that $g$ is continuously differentiable and has either compact support or has sufficient decay, which guarantees the well-posedness of \eqref{evolution_ODE}. In addition, we assume that $g$ is a positive definite function, which leads to a specific representation (see Section~\ref{sec:radial}) that is used to derive the proposed model although these assumption will be relaxed for applications. For simplicity, denote $\mathbf{x}(t):= (\mathbf{x}_1(t),\dots,\mathbf{x}_n(t))$ and  $\mathbf{r}_{:,i}(t) =  (\mathbf{x}_1(t)-\mathbf{x}_i(t),\dots,\mathbf{x}_n(t)-\mathbf{x}_i(t))$. Equation~\eqref{evolution_ODE} can be derived from the gradient flow of the potential energy $\mathcal{U}(\mathbf{x}(t))=\sum_{i\not=i'} G(\mathbf{r}_{i',i}(t))$ where $g(r) := \frac{G'(r)}{r}$. Such gradient flow systems have proven effective as a minimal framework to model a variety of chemical and living systems such as for micelle formation \cite{somoza1995model} or aggregation and swarming of living things, even without the addition of stochastic noise.  The pairwise interaction kernel can be derived from physical constraints for the potential energy in non-living systems, but this type of first principles approach becomes unreasonable to derive the governing equations for biological systems. 

\subsection{New Random Features}\label{sec:radial}
Motivated by the characterization of radial positive definite functions on $\mathbb{R}^d$ from \cite{schoenberg1938metric} and the earlier work on spherical random feature models \cite{pennington2015spherical}, we propose approximating the interacting kernel $g$ in \eqref{evolution_ODE} using a random radial feature space defined by Gaussians centered at zero. 
\begin{thm}[\cite{schoenberg1938metric}] \label{thm:PDF}
A continuous function $g:\mathbb{R}_+\to \mathbb{R}$ is radial and positive definite on $\mathbb{R}^d$ for all $d$ if and only if it is of the form $g(r) = \int_0^\infty e^{-r^2\omega^2}\;d\nu(\omega)$ where $\nu$ is a finite non-negative Borel measure on $\mathbb{R}_+$.
\end{thm}

We consider a weighted integral representation, namely, we assume that  $g(r) = \int_0^\infty \alpha(\omega) e^{-r^2\omega^2}\;d\nu(\omega)$, then \eqref{evolution_ODE} becomes
\begin{align}\label{eq:RF_ode}
   & \frac{d}{dt} {\mathbf x}_i(t) = \frac{1}{n} \sum_{i'=1}^n\limits g(\Vert    \mathbf{r}_{i',i}(t)\Vert_2)\, \mathbf{r}_{i',i}(t) \notag \\
&= \frac{1}{n} \left(\sum_{i'=1}^n \int_0^\infty \alpha(\omega) e^{-\|\mathbf{r}_{i',i}(t)\|_2^2\, \omega^2}\;d\nu(\omega)\right)\mathbf{r}_{i',i}(t)\notag \\
&= \int_0^\infty  \left(\frac{\alpha(\omega)}{n} \sum\limits_{i'=1}^n e^{- \|\mathbf{r}_{i',i}(t)\|_2^2\, \omega^2}\, \mathbf{r}_{i',i}(t) \right) \;d\nu(\omega) \notag \\
&= \int_0^\infty  \phi( \mathbf{r}_{1,i}(t), \ldots, \mathbf{r}_{n,i}(t), \omega) \;d\nu(\omega)\\
&=: F( \mathbf{r}_{1,i}(t), \ldots, \mathbf{r}_{n,i}(t)),\notag
\end{align}
for all $i \in [n]:=\{1, \ldots, n\}$. In \eqref{eq:RF_ode}, the integrand $\phi$ is parameterized by the scalar $\omega\in \mathbb{R}_+$ with respect to the measure $\nu$. To approximate \eqref{eq:RF_ode}, we build a set of $N$ random features by sampling $\omega$ from a user defined probability measure $\theta$, i.e. $\omega_k \sim \theta(\omega)$ for $k \in [N]$.  The function $F$ does not explicitly depend on the agents' locations or time and instead is a function of the radial distances. Thus for simplicity consider $F( \mathbf{r}_{1}, \ldots, \mathbf{r}_{n})$, i.e. a function of $n$ inputs, then the random feature approximation for this problem becomes 
\begin{align}\label{eq:RF_F}
F_N( \mathbf{r}_{1}, \ldots, \mathbf{r}_{n}) &= \frac{1}{N} \sum\limits_{k=1}^N c_k\, \phi_k( \mathbf{r}_{1}, \ldots, \mathbf{r}_{n}) 
\end{align}
where $\mathbf{c}=(c_1, \ldots, c_N)$ are the coefficients of the expansion of $F_N$ with respect to the proposed radial feature space 
\begin{align*}
\left\{\phi_k\, \bigg|\, \phi_k( \mathbf{r}_{1}, \ldots, \mathbf{r}_{n})= \frac{\alpha(\omega_k)}{n} \sum\limits_{i=1}^n  e^{- \|\mathbf{r}_i\|_2^2\, \omega_k^2}\, \mathbf{r}_{i}\right\}.
\end{align*}
The weight $\alpha(\omega)$ is chosen to normalize the basis functions $\phi_k$, i.e. we set
\begin{equation}\label{eq:normal}
\alpha(\omega)=\left(\int_0^\infty  \frac{1}{n} \sum\limits_{i=1}^n e^{- \|\mathbf{\mathbf{r}} \|_2^2\, \omega^2}\, r_{j} \, \;dr_{1} \ldots \;dr_{n} \right)^{-1}=\frac{2^n\omega^{n+1}}{\pi^{\frac{n-1}{2}}}.
\end{equation}
The motivation for rescaling the terms is to avoid the ill-conditioning which occurs for larger $\omega_k$, that is, the Gaussians become nearly zero which leads to poor generalization. We observed that rescaling by an increasing function of $\omega$, e.g. $\omega^p$ for $p\geq 1$ was sufficient to obtain meaningful results; however, for theoretical consistency we use the normalization factor \eqref{eq:normal}. Although the random feature representation for the interaction kernel $g$ is similar to \cite{pennington2015spherical}, the approximation to $F$ differs since we are considering a vectorized system of $n$-agents and a different functional form.   

\begin{figure}[b!]
\centering
\includegraphics[scale=0.3]{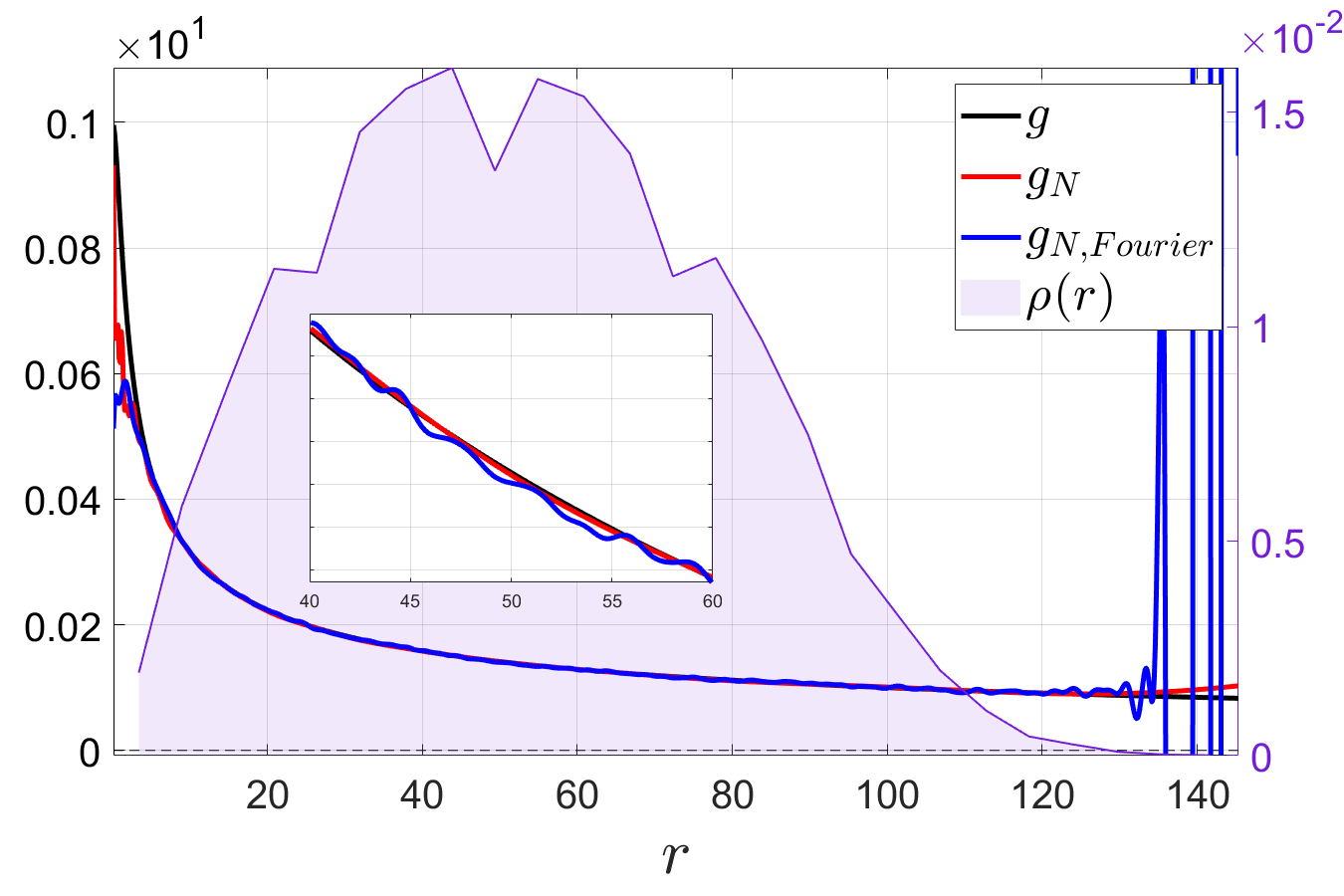}
    \caption{Radial Features versus Fourier Features: The plot includes the true interaction kernel (the black curve) and the learned interaction kernels using the radial features defined by  \eqref{eq:RF_F} (the red curve) and the random Fourier features (blue curve). The purple area displays the empirical distribution of the radial distances estimated on the trajectories.}
    \label{fig:compare_with_Fourier}
\end{figure}

In Figure \ref{fig:compare_with_Fourier}, we compare the formulation \eqref{eq:RF_F} with random Fourier features (RFF) \cite{rahimi2007random, rahimi2008uniform, rahimi2008weighted}.  The global and oscillatory nature of RFF does not match the typical behavior of the interaction kernels, i.e. the specific growth and decay properties either near the origin or in far-field. This leads to a smooth but oscillatory approximation of the kernel that causes a large accumulation of errors when forecasting the states using the learned dynamical system. In the zoomed-in plot (for region $r\in[40,60]$), we see that the RFF produces a non-monotone and ``noisy'' fit. In this example, we see that our approach outperforms the RFF by matching the structural assumptions. 

\subsection{Least Squares Learning Problem}

The training problem is to approximate $F$ in \eqref{eq:RF_ode} by $F_N$ from \eqref{eq:RF_F} given a set of discrete trajectory paths. That is, given  $J$ observation timestamps $\{ t_j \}_{j\in[J]}$ with $t_j \in [0,T]$ and $t_{j}<t_{j+1}$ for all $j \in [J]$ and a total of $L$ initial conditions (IC) $\mathbf{x}^\ell(0) = (\mathbf{x}_1^\ell(0),\dots,\mathbf{x}_n^\ell(0))$, we observe $L$ trajectories indexed by $\ell \in [L]$ and stored as $ \mathbf{X}^{\ell}=[\mathbf{x}^\ell(t_j)]_{j \in [J], i \in [n]}\in \mathbb{R}^{dnJ}$. The entire training set is the concatenation of the $L$ trajectories 
$\mathbf{X}=[\mathbf{X}^{\ell}]_{\ell \in [L]} \in \mathbb{R}^{n_\text{tot}}$,
where $n_\text{tot} = dnJL$, i.e. the product between the dimension of each agent, the number of agents, the number of timestamps, and the total number of trajectories. The IC $\mathbf{x}^\ell(0)$ are drawn independently from a prescribed probability measure $\mathbf{\mu}_0$ on $\mathbb{R}^{dn}$. The velocity can be estimated by a finite difference method and concatenated to form the velocity dataset $\mathbf{V}=[\mathbf{V}^{\ell}_{i,j}]_{\ell \in [L]} \in \mathbb{R}^{n_\text{tot}}$,
where $\mathbf{V}^{\ell}$ is an approximation to $\frac{d}{dt} \mathbf{x}_{i}^\ell(t_j)$. We used the central difference formula in all examples unless otherwise stated.
 
To train the system, we want to minimize the risk with respect to the unknown non-negative Borel measure $\nu$:
\begin{align*}
    \mathcal{L}(\nu) := \frac{1}{nLT}\sum\limits_{i \in[n], \ell \in[L]} \int_0^T \norm{\frac{d}{dt} \mathbf{x}_{i}^\ell(t) - F( \mathbf{r}^\ell_{:,i}(t); \nu)}_2^2 \text{dt},  
\end{align*}
which measures the error between the velocity and the RHS of the ODE \eqref{eq:RF_ode} where $F( \cdot\, ; \nu)$ is used to denote the dependence of $F$ on $\nu$. This is referred to as a \textit{trajectory based regression} problem. The empirical risk is given by 
\begin{equation} \label{eq:emp_risk}
\mathcal{L}_N(\mathbf{c})  := \frac{1}{nJL}\sum\limits_{i \in[n], j \in [J], \ell \in[L]} \left|\mathbf{V}_{i,j}^\ell - F_N( \mathbf{r}^{\ell}_{:,i}(t_j); \mathbf{c}) \right|^2,
\end{equation}
which measures the error between the estimated (or observed) velocity and the random feature model $F_N( \cdot \, ; \mathbf{c})$ from \eqref{eq:RF_F} (which depends on the vector $c$). The sum in \eqref{eq:emp_risk} is evaluated on a finite set of timestamps. 
To simplify the expression for $F_N$, let $\mathbf{A} \in \R^{n_\text{tot}\times N}$ be the random radial feature matrix whose elements are defined by $a_{\tilde{i},k} = \phi_k(\textbf{r}^\ell_{:,i}(t_j))$ where $\tilde{i}$ represents the triple $(i,j,\ell)$ after reindexing to a single index. Then \eqref{eq:emp_risk} can be written as
\begin{equation} \label{eq:emp_risk_vect}
\mathcal{L}_N(\mathbf{c}) = \frac{1}{nJL} \norm{\mathbf{V} - \mathbf{A} \mathbf{c}}_2^2,   
\end{equation}
using vector notation. 
We can obtain the trained coefficient vector by minimizing the empirical loss directly, i.e.
\begin{equation} \label{eq:OptL2}
\mathbf{c}= \arg\min_{\mathbf{c}' \in \mathbb{R}^N}\, \norm{\mathbf{V} - \mathbf{A} \mathbf{c}'}_2^2,   
\end{equation}
i.e. the ordinary least squares problem.

\begin{figure}[h!]
\centering
\includegraphics[scale=0.22]{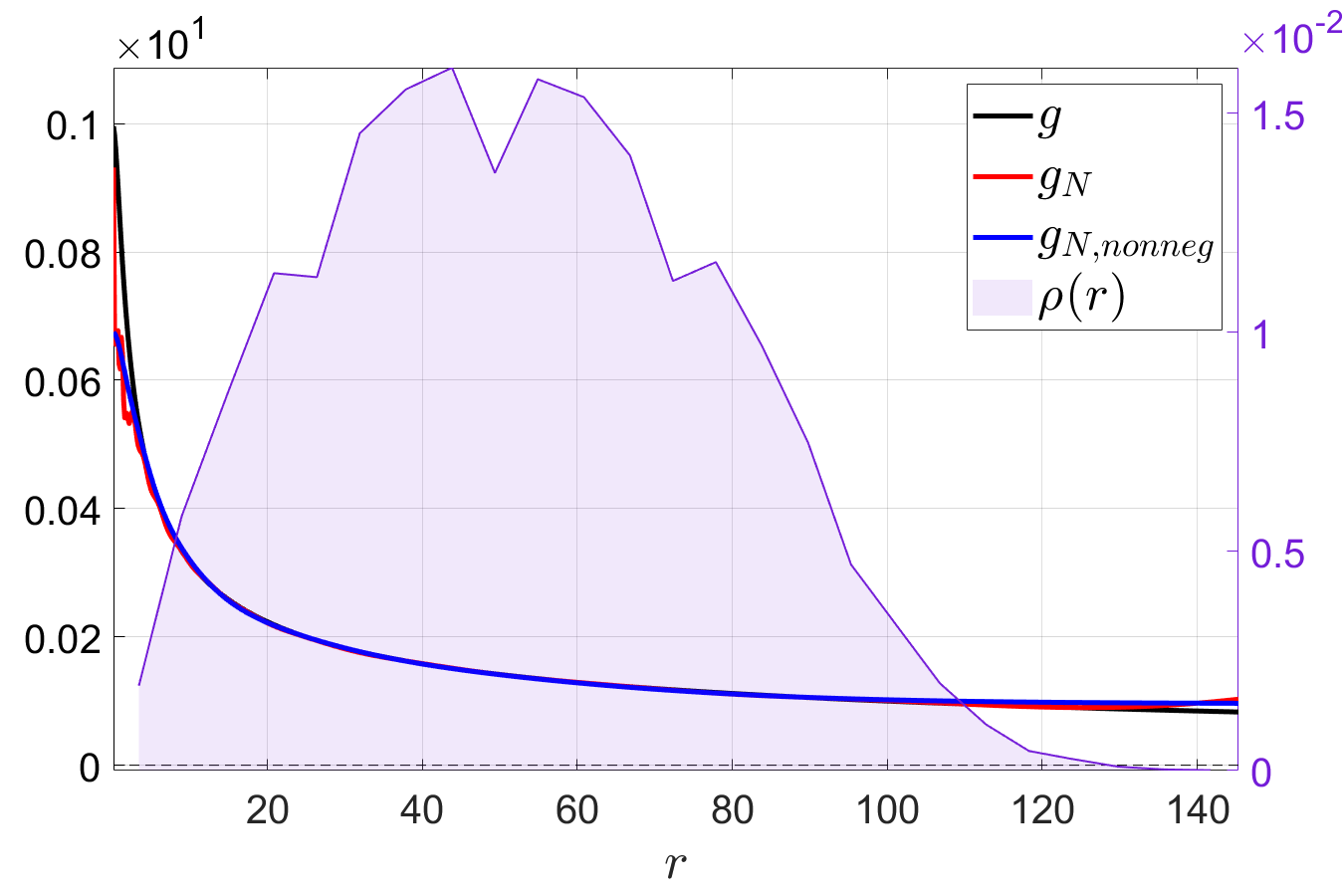}
 \includegraphics[scale = 0.22]{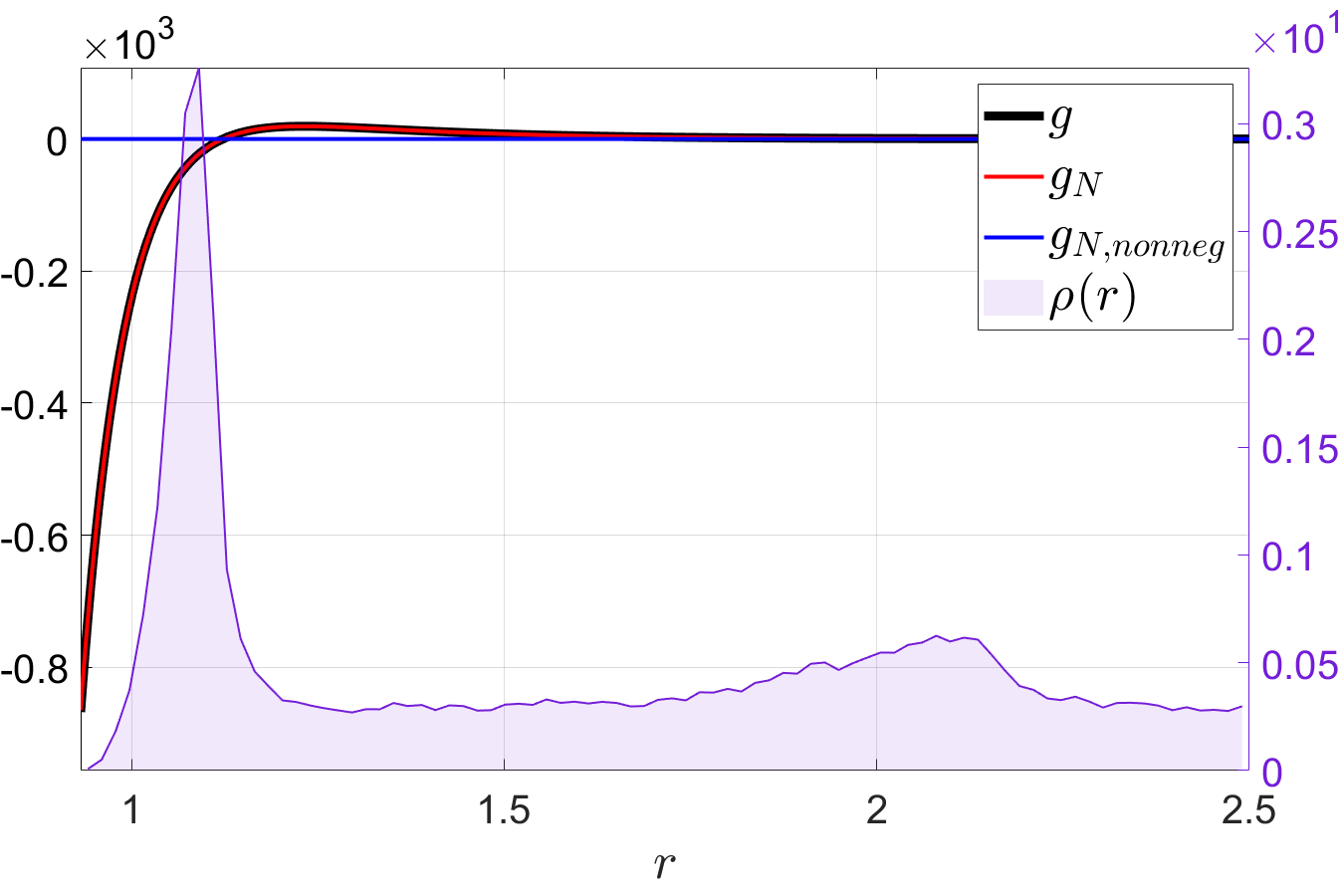}
    \caption{Comparing Constraints: The plots compare the true interaction kernels (the black curves) and the learned interaction kernels using the radial features with the non-negative constraints (the red curves) and with the non-negative constraints (the blue curves). The purple region plots the empirical distribution of the radial distances estimated on the trajectories. The first graph shows that (within the region of observed radial distances), the two kernels obtain similar accuracy when the true kernel satisfies the constraints. The second graph shows that if the true kernel does not satisfy the constraints, then the unconstrained approach is preferred.}
    \label{fig:compare_constraint}
\end{figure}

Note that, for a positive definite function, the coefficients $\mathbf{c}$ should be constrained to be non-negative based on Theorem~\ref{thm:PDF}. However, by relaxing the assumptions we find that the approximation is still valid in practice without the additional computational cost incurred by adding a constraint. In Figure~\ref{fig:compare_constraint}, we compare the approximation of two interaction kernels (the true kernels are the black curves) using the unconstrained optimization formulation (in red) and the non-negative constrained optimization formulation (in blue). The (empirical) density function of radial distances is represented by the purple region in all figures.  The first plot in Figure~\ref{fig:compare_constraint} shows that for a radial and positive definite kernel $g$, the non-negative constrained problem produces a smoother approximation that agrees better with the true kernel for small $r>0$. However, the regions where $g_N$ does not agree with $g$ have low sampling density and are thus less likely events in the dynamics. The second plot in Figure~\ref{fig:compare_constraint} provides an example where the true interaction kernel $g$ does not satisfy the assumptions of Theorem \ref{thm:PDF}. In this case, the kernel $g(r)\rightarrow -\infty$ as $r\rightarrow 0^+$, thus for visualization we do not include the large (negative) values. The non-negative constrained solutions is trivial since positive coefficients add more deviation for $0<r<1$ than the potential fit gained for the region $r>1$.  Based on this example, we argue that for positive definite and radial kernels the discrepancies between the two formulations are not statistically significant, while for non-positive definite kernels the non-negative constraint can lead to large errors. Thus without prior knowledge about the system, we use the unconstrained formulation.

The training problem \eqref{eq:emp_risk_vect} measures the  ``stationary'' error,  since it compares the data and model using the observed velocity at each point in time. Another way to state this is that the features, i.e. the columns of $\mathbf{A}$, are applied to the data directly and do not depend on the learned solution generated by the trained system. 
To evaluate the effectiveness of the trained model, we want to measure the error of the model (i.e. the equations of motion) and the error that is incurred over the path generated by the trained ODE system: 
\begin{align} \label{eq:trained_ODE}
\begin{cases}
    &\frac{d}{dt} \tilde{\mathbf{x}}_i(t) = F_N( \tilde{\mathbf{r}}_{1,i}(t), \ldots, \tilde{\mathbf{r}}_{n,i}(t); c)\\
    &\tilde{\mathbf{r}}_{i',i}(t) =  \tilde{\mathbf{x}}_{i'}(t)-\tilde{\mathbf{x}}_i(t).
    \end{cases}
\end{align}
To define the generalization error of the model, consider the probability measure $\rho$ on $\mathbb{R}_+$ that defines the distribution of pairwise distances between agents. This gives us the needed distribution for the values of $r$ that define the input distribution to the true interaction kernel $g$ and the trained interaction kernel defined by
\begin{align}\label{eq:RF_g}
g_N({\mathbf r}) = \frac{2^n}{\pi^{\frac{n-1}{2}}\, N} \sum\limits_{k=1}^N c_k\, \omega_k^{n+1} \,e^{- \|{\mathbf r}\|_2^2\, \omega_k^2}.
\end{align}
Formally, for any interval $U\subseteq \mathbb{R}_+$, $\rho(U)$ is the probability that the pairwise distance between two agents is in $U$, for all IC sampled from $\mathbf{\mu}_0$ and for $t\in[0,T]$, i.e. 
\begin{equation}\label{eq:density_of_radius}
    \rho(r) = \frac{1}{\binom{n}{2}T}\int_{0}^T \mathbb{E}_{\mathbf{x}(0)\sim \mathbf{\mu}_0}\left[\sum\limits_{1\leq i'<i \leq n} \delta_{r_{i',i}(t)}(r) \right]\;dt.
\end{equation}
We define the $L^2(\rho)$ \textit{kernel generalization error} as $\| G'-G'_N\|_{L^2(\rho)}$ where $G'(r) = g(r) r$ and $G_N'(r) = g_N(r) r$. In practice, the expectation and supremum cannot be computed directly, so instead one can consider the empirical probability density by replacing the integrals and expectations via averages. 
Lastly, the \textit{path-wise generalization error} can be defined as
\begin{equation}\label{eq:path_error}
     \mathbf{E}(c)=\E_{\mathbf{x}(0)\sim \mathbf{\mu}_0}\left[\sup_{t\in [0,\tilde{T}],i\in [N]} \norm{\mathbf{x}_i(t) - \tilde{\mathbf{x}}_i(t) }_2\right]
\end{equation}
where $\tilde{\mathbf{x}}$ is the simulated path using \eqref{eq:trained_ODE} and thus depends implicitly on $c$. When final forecast time $\tilde{T}=T$,  $\mathbf{E}$ measures how close the simulated path is to the training data, noting that this is not the training error used in \eqref{eq:emp_risk_vect} since the training problem minimizes the stationary risk and not the discrepancy between the data and the simulated trajectory. For all experiments, we use $\tilde{T}>T$ to evaluate the performance of the model in the extrapolation regime, i.e. how well does the model predict future states using a test dataset. The error \eqref{eq:path_error} is approximated numerically by averaging over a finite set of randomly sampled paths and maximized over a finite set of timestamps.

The errors depend on the ability of the training problem to approximate the interaction kernel $g$. For functions in the appropriate reproducing kernel Hilbert spaces,  it was shown that under some technical assumptions, a random feature approximation obtained from minimizing the empirical risk with a ridge regression penalty (i.e. adding the term $\lambda \|\mathbf{c}\|_2^2$ to \eqref{eq:emp_risk_vect})   produces generalization errors that scale like $\mathcal{O}(m^{-\frac{1}{2}})$ when the number of random features scales like $N=\mathcal{O}({m}^{\frac{1}{2}} \, \log{m})$ where $m$ is the total number of data samples \cite{rudi2017generalization}. For the ordinary least squares case, \cite{mei2021generalization, chen2021conditioning} analyzed the behavior of the approximation and its associated risk as a function of the dimension of the data, the size of the training set, and the number of random features. In the overparameterized setting, i.e. when there are many more features than data samples, \cite{mei2021generalization} concluded that the ordinary least squares model may produce the optimal approximation amongst all kernel methods. In \cite{chen2021conditioning,chen2022concentration}, the random feature matrix was shown to be well-conditioned in the sense that its singular values were bounded away from zero with high probability. In addition, the risk associated with underparameterized random feature methods scale like $\mathcal{O}(N^{-1}+m^{-\frac{1}{2}})$. Although the assumptions in \cite{rudi2017generalization, mei2021generalization,chen2021conditioning} do not directly hold in our setting, their results do indicate the expected behavior of using a RFM for learning interacting dynamics in the regime where the number of agents or the number of random initial conditions is sufficiently large (thus matching the random sampling needed in the theory for random feature methods). We leave a detailed theoretical examination of this problem for future work.

\subsection{Sparse Learning Problem}
The theory for RFMs suggests that $N$ must be sufficiently large to obtain high accuracy models. However, the cost of simulating the path using \eqref{eq:trained_ODE} requires at least one query of the random feature model for each time-step and thus scales linearly in $N$. Therefore, for computational efficiency one does not want $N$ to be too large but for accuracy large $N$ is needed. This motivates us to use a sparse random feature regression problem, in particular, we replace \eqref{eq:OptL2} with the $\ell_0$ constrained least squares problem:
\begin{equation} \label{eq:OptL0}
{\mathbf c}= \arg\min_{\substack{\mathbf{c}' \in \mathbb{R}^N, \|\mathbf{c}'\|_0 \leq s}}\ \norm{\mathbf{V} - \mathbf{A} \mathbf{c}}_2^2,   
\end{equation}
where $\|\mathbf{c}\|_0$ is the number of nonzero entries of $\mathbf{c}$. To solve \eqref{eq:OptL0}, we use a greedy approach called the hard thresholding pursuit (HTP) algorithm \cite{foucart2011hard}
\begin{align}
\begin{cases}
    & S^{h+1} =  \{ \text{indices of }s\  \text{largest (in magnitude) entries of} \\
    & \hspace{2cm} \mathbf{c}^h + \mathbf{A}^*(\mathbf{V} - \mathbf{A}\mathbf{c}^h) \}\\
     & \mathbf{c}^{h+1} = \arg \min\{ \| \mathbf{V} - \mathbf{A}\mathbf{c}\|_2^2,\quad  \text{supp}(\mathbf{c})\subseteq S^{h+1}\}
 \end{cases} \label{eq: HTP1}
\end{align}
which generates a sequence $\mathbf{c}^{h}$ index by $h\in[0,H]$. The two step approach iteratively sparsifies and then re-fits the coefficients with respect to the random radial feature matrix $\mathbf{A}$ applied to the training dataset. 
 
\begin{figure}[t!]
    \centering
    \includegraphics[scale=0.65]{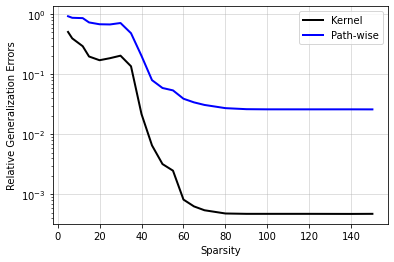}
    \caption{Motivation for Sparsity: The plot shows the relative generalization errors for the kernel and learned trajectories as a function of the coefficient sparsity level $s$ for the Lennard-Jones system. The errors plateau for $s\geq 80$ indicating that adding more features does not necessarily improve the error in this example.}
    \label{fig:sparsity_graph}
\end{figure}

The motivation for using sparsity is that it enables us to find the most dominate modes from a large feature space. Theoretically, we expect that the sparse coefficients $\mathbf{c}$ concentrate near the largest values of $\nu(\omega)$ when taking into account the random sampling density $\theta(\omega)$. To verify the sparsity assumption, we measure the  (relative) generalization errors produced by the random radial basis approximation obtain from the HTP algorithm \eqref{eq: HTP1} with various sparsity levels $s$ in the range $[1,150]$; see Figure~\ref{fig:sparsity_graph}. Both relative errors generally decay as $s$ increases and we see that they plateau for $s\geq 80$, thus no accuracy gains are made by increasing the number of terms used in the approximation. When $s$ is close to $40$, the errors are within an order of magnitude of the optimal error. Thus, it would be advantageous to choose a sparsity level $s$ smaller than the dimension of the full feature space in order to lower the simulation cost of \eqref{eq:trained_ODE}.

There have been several recent random feature methods that either use an $\ell^1$ or an $\ell^0$ penalty to obtain sparse representations. In \cite{hashemi2021generalization}, the sparse random feature expansion was proposed which utilized an $\ell^1$ basis pursuit denoising formulation in order to obtain sparse coefficients with statistical guarantees, see also  \cite{chen2021conditioning}.  The results in \cite{hashemi2021generalization} match the risk bounds of the $\ell^2$ formulation, i.e. an error of $\mathcal{O}\left(N^{-1} + m^{-\frac{1}{2}}\right)$, when the sparsity level is set to $s=N$ and the number of measurements $m= \mathcal{O}(N \log(N))$. When the coefficient vector is \textit{compressible}, i.e. they can be well-represented by a sparse vector, then $s<N$ gives similar results with a lower model complexity than in the $\ell^2$ case and with fewer samples $m= \mathcal{O}(s \log(N))$. In \cite{yen2014sparse}, a step-wise LASSO approach was used to iteratively increase the number of random features by adding a sparse subset from multiple random samples.  In \cite{xie2021shrimp}, an iterative magnitude pruning approach was used to extract sparse representations for RFM. Our proposed model is related to the hard-ridge random feature model \cite{saha2022harfe} where it was shown that for certain feature functions $\phi_k$ with Gaussian data and weights, the iterative method converges to a solution with a given error bound, see also \cite{richardson2022srmd}.

Although both the $\ell^2$ and the sparse random feature models come with various theoretical properties, those results do not hold directly here. In the setting of learning multi-agent interaction kernels, the data samples are typically less ideal than in the theoretical setting, namely, trajectories are highly correlated and so they are not independent and identically distributed random variables. In fact, the randomness in the data is obtained from the sampling of various initial conditions.  Additionally, the trajectories may contain noise and outliers which leads to non-standard biasing in the velocity estimates in the training problem. Empirically, we observe that the two approaches behave in a similar fashion, and that the sparsity-promoting approach often avoids overfitting on the data.

\section{Numerical Method}

The datasets are created by approximating the solutions to the systems of ODE (see Section~\ref{sec:examples}) in MATLAB using the ODE15s solver with relative tolerance $10^{-5}$ and absolute tolerance $10^{-6}$ to handle the stiffness of the system. The train and test datasets have the same size collection of $L$ independent trajectories computed over the time interval $[0,T]$. The $L$ initial conditions are independently sampled from the user prescribed probability measure $\mathbf{\mu}_0$, i.e. $\mathbf{x}^{\ell}(0) \sim \mathbf{\mu}_0$ for $\ell \in [L]$ which is specified in each example. The observations are the state variables (i.e. we have measurements of $\mathbf{x}(t)$ and we do not have direct measurements of the velocity $\frac{d}{dt}\mathbf{x}(t)$) with multiplicative uniform noise of magnitude $\sigma_{\text{noise}}$ at $J$ equidistant time-stamps in the interval $[0,T]$.  An approximation of the unknown distribution $\rho(r)$ (defined in \eqref{eq:density_of_radius}) is obtained by using a large number $L'$ ($L'\gg L$) of independent trajectories and averaging these trials to estimate $\rho(r)$ empirically.  This is used for visualizing the distribution of the data and assessing the error of the learned kernel.

\subsection{Loss Function for First-order Homogeneous Systems}
Given $n$ agents $\{\mathbf{x}_i(t)\}_{i=1}^n,\mathbf{x}_i(t)\in \R^d$ for all $i\in [n]$, the first-order homogeneous system is governed by the following system of ODEs: \begin{align*} 
\begin{cases}
    &\frac{d}{dt} {\mathbf x}_i(t) = \frac{1}{n} \sum\limits_{i'=1}^n g(\Vert    \mathbf{r}_{i',i}(t)\Vert_2)\, \mathbf{r}_{i',i}(t) =: F(\mathbf{r}_{1,i}(t),\dots,\mathbf{r}_{n,i}(t))\\
    &\mathbf{r}_{i',i}(t) =  \mathbf{x}_{i'}(t)-\mathbf{x}_i(t)
    \end{cases}
\end{align*} where $g$ is the interaction kernel. Consider the following radial feature space 
\begin{align*}
\left\{\phi_k\, \bigg|\, \phi_k( \mathbf{r}_{1}, \ldots, \mathbf{r}_{n})= \frac{\alpha(\omega_k)}{n} \sum\limits_{i=1}^n  e^{- \|\mathbf{r}_i\|_2^2\, \omega_k^2}\, \mathbf{r}_{i}\right\}.
\end{align*}
We seek to approximate $F$ using a linear combination of elements from the above radial feature space: 
\begin{align*}
    F_N( \mathbf{r}_{1}, \ldots, \mathbf{r}_{n}) &= \frac{1}{N} \sum\limits_{k=1}^N c_k\, \phi_k( \mathbf{r}_{1}, \ldots, \mathbf{r}_{n}).
\end{align*}
Denote the (approximated) velocity dataset by $\mathbf{V} =[\mathbf{V}_{i,j}^\ell]_{\ell\in [L]}= \left[\frac{d}{dt}\mathbf{x}_i^\ell(t_j)\right]\in \R^{n_\text{tot}}$ and the radial feature matrix for space $\{\phi_k\st k\in [N]\}$ by $\mathbf{A} = [a_{\tilde{i},k}] =  [\phi_k(\textbf{r}^\ell_{:,i}(t_j))]\in \R^{n_\text{tot}\times N}$ where $\tilde{i}$  where $\tilde{i}$ represents the triple $(i,j,\ell)$ after reindexing to a single index, the loss function is \begin{align*}
    \mathcal{L}_N(\mathbf{c})  &= \frac{1}{nJL}\sum\limits_{i \in[n], j \in [J], \ell \in[L]} \left|\mathbf{V}_{i,j}^\ell - F_N( \mathbf{r}^{\ell}_{:,i}(t_j); \mathbf{c}) \right|^2\\
    &= \frac{1}{nJL} \norm{\mathbf{V} - \mathbf{A} \mathbf{c}}_2^2.
\end{align*}

\subsection{Loss Function for First-order Heterogeneous Systems}
In a first-order heterogeneous system, each of the $n$ agents $\{\mathbf{x}_i(t)\}_{i=1}^n$ has a type label $k_i\in \{1,2\}$. We use $n_1$ and $n_2$ to denote the number of agents of type $1$ and $2$. The system is governed by the following system of ODEs: \begin{align*} 
\begin{cases}
    &\frac{d}{dt} {\mathbf x}_i(t) = \sum\limits_{i'=1}^n \frac{1}{n_{k_{i'}}} g_{k_ik_{i'}}(\Vert    \mathbf{r}_{i',i}(t)\Vert_2)\, \mathbf{r}_{i',i}(t) =: F^\text{heterog.}(\mathbf{r}_{:,i}(t))\\
    &\mathbf{r}_{i',i}(t) =  \mathbf{x}_{i'}(t)-\mathbf{x}_i(t)
\end{cases}
\end{align*} where $g_{k_ik_{i'}}$ is the interaction kernel governing how agents of type $k_{i'}$ influence agents of type $k_i$. Consider the following radial feature space 
\begin{align*}
\left\{\phi_k^\text{heterog.}\, \bigg|\, \phi_k^\text{heterog.}( \mathbf{r}_{1}, \ldots, \mathbf{r}_{n})=  \sum\limits_{i=1}^n \frac{\beta(\omega_k)}{n_{k_{i}}}  e^{- \|\mathbf{r}_i\|_2^2\, \omega_k^2}\, \mathbf{r}_{i}\right\}.
\end{align*}
We seek to approximate $F^\text{heterog.}$ using a linear combination of elements from the above radial feature space:
\begin{align*}
F_N^\text{heterog.}( \mathbf{r}_{1}, \ldots, \mathbf{r}_{n}) &= \frac{1}{N} \sum\limits_{k=1}^N c_k\, \phi_k^\text{heterog.}( \mathbf{r}_{1}, \ldots, \mathbf{r}_{n})
\end{align*}
where $\beta(\omega_k) = \frac{\sqrt{\pi}}{2\omega_k}$ is a normalization factor.
Denote the (approximated) velocity dataset by $\mathbf{V} =[\mathbf{V}_{i,j}^\ell]_{\ell\in [L]}= \left[\frac{d}{dt}\mathbf{x}_i^\ell(t_j)\right]\in \R^{n_\text{tot}}$ and the radial feature matrix for space $\{\phi_k^\text{heterog.}\st k\in [N]\}$ by $\mathbf{A} = [a_{\tilde{i},k}] =  [\phi_k^\text{heterog.}(\textbf{r}^\ell_{:,i}(t_j))]\in \R^{n_\text{tot}\times N}$, the loss function is \begin{align*}
    \mathcal{L}_N^\text{heterog.}(\mathbf{c})  &= \frac{1}{nJL}\sum\limits_{i \in[n], j \in [J], \ell \in[L]} \left|\mathbf{V}_{i,j}^\ell - F_N^\text{heterog.}( \mathbf{r}^{\ell}_{:,i}(t_j); \mathbf{c}) \right|^2\\
    &= \frac{1}{nJL} \norm{\mathbf{V} - \mathbf{A} \mathbf{c}}_2^2.
\end{align*}

\subsection{Loss Function for Second Order Homogeneous Systems}
Given $n$ agents $\{\mathbf{x}_i(t)\}_{i=1}^n,\mathbf{x}_i(t)\in \R^d$ for all $i\in [n]$, the second order homogeneous system is governed by the following system of ODEs: \begin{align*}
\begin{cases}
    &\frac{d^2}{dt^2} {\mathbf x}_i(t) = \frac{1}{n} \sum\limits_{i'=1}^n g(\Vert    \mathbf{r}_{i',i}(t)\Vert_2)\, \frac{d}{dt}\mathbf{r}_{i',i}(t)=:F^\text{sec.}\left(\mathbf{r}_{:,i}(t),\frac{d}{dt}\mathbf{r}_{:,i}(t)\right)\\
    &\mathbf{r}_{i',i}(t) =  \mathbf{x}_{i'}(t)-\mathbf{x}_i(t)
    \end{cases}
\end{align*} where $g$ is the interaction kernel. 
Consider the following radial feature space 
\begin{align*}
 \left\{\phi_k^\text{sec.}\, \bigg|\, \phi_k^\text{sec.}( \mathbf{r}_{1}, \ldots, \mathbf{r}_{n})= \frac{\beta(\omega_k)}{n} \sum\limits_{i=1}^n  e^{- \|\mathbf{r}_i\|_2^2\, \omega_k^2}\, \frac{d}{dt}\mathbf{r}_{i}\right\}.
 \end{align*}
We seek to approximate $F^\text{sec.}$ using a linear combination of elements from the above radial feature space:
\begin{align*}
F_N^\text{sec.}\left( \mathbf{r}_{:},\frac{d}{dt}\mathbf{r}_{:,i}(t)\right)= \frac{1}{N} \sum\limits_{k=1}^N c_k\, \phi_k^\text{sec.}\left( \mathbf{r}_{:},\frac{d}{dt}\mathbf{r}_{:,i}(t)\right).
\end{align*}
Denote the (approximated) acceleration dataset by $\mathbf{V} =[\mathbf{V}_{i,j}^\ell]_{\ell\in [L]}= \left[\frac{d^2}{dt^2}\mathbf{x}_i^\ell(t_j)\right]\in \R^{n_\text{tot}}$ and the radial feature matrix for space $\{\phi_k^\text{sec.}\st k\in [N]\}$ by $\mathbf{A} = [a_{\tilde{i},k}] =  \left[\phi_k^\text{sec.}\left(\mathbf{r}^\ell_{:,i}(t_j),\frac{d}{dt}\mathbf{r}^\ell_{:,i}(t_j)\right)\right]\in \R^{n_\text{tot}\times N}$, the loss function is \begin{align*}
    \mathcal{L}_N^\text{sec.}(\mathbf{c})  &= \frac{1}{nJL}\sum\limits_{i \in[n], j \in [J], \ell \in[L]} \left|\mathbf{V}_{i,j}^\ell - F_N^\text{sec.}\left( \mathbf{r}^{\ell}_{:,i}(t_j),\frac{d}{dt}\mathbf{r}^{\ell}_{:,i}(t_j); \mathbf{c}\right) \right|^2\\
    &= \frac{1}{nJL} \norm{\mathbf{V} - \mathbf{A} \mathbf{c}}_2^2.
\end{align*}

\subsection{Remarks}
Given training set $\mathbf{X}^\ell = [\mathbf{x}^\ell(t_j)]_{j\in [J],i\in [n]}\in \R^{dnJ}$ for $\ell\in [L]$, we first obtain the velocity dataset $\mathbf{V} =[\mathbf{V}_{i,j}^\ell]_{\ell\in [L]}= [\frac{d}{dt}\mathbf{x}_i^\ell(t_j)]\in \R^\text{tot}$ using the central difference scheme: \begin{equation*}
    \frac{d}{dt}\mathbf{x}_i^\ell(t_j)\approx  \frac{\mathbf{x}_i^\ell(t_{j+1})-\mathbf{x}_i^\ell(t_{j-1})}{t_{j+1}-t_{j-1}},
\end{equation*}
and we do not assume that the samples are evenly spaced in time.
After having the state and (approximated) velocity information as training data, we randomly sample $\omega_k$ from $\n(0,\sigma^2)$ (where $\sigma^2$ is user-defined) to form the radial feature space for learning: \begin{align*}
\left\{\phi_k\, \bigg|\, \phi_k( \mathbf{r}_{1}, \ldots, \mathbf{r}_{n})= \frac{\alpha(\omega_k)}{n} \sum\limits_{i=1}^n  e^{- \|\mathbf{r}_i\|_2^2\, \omega_k^2}\, \mathbf{r}_{i}\right\}.
\end{align*} 
We then assemble them into a random radial feature matrix $\mathbf{A}\in \R^{n_\text{tot}\times N}$ whose elements are defined by $a_{\tilde{i},k} = \phi_k(\textbf{r}^l_{:,i}(t_j))$. The empirical loss $\mathcal{L}_N(\mathbf{c}) = \frac{1}{nJL} \norm{\mathbf{V} - \mathbf{A} \mathbf{c}}_2^2$ is minimized using either least squares or HTP with sparsity $s$ to obtain coefficients $\mathbf{c}$, and we form the approximations $g_N\approx g$ and $F_N\approx F$ via 
\begin{align*}
g_N({\mathbf r}) &= \frac{2^n}{\pi^{\frac{n-1}{2}}\, N} \sum\limits_{k=1}^N c_k\, \omega_k^{n+1} \,e^{- \|{\mathbf r}\|_2^2\, \omega_k^2}\\
    F_N( \mathbf{r}_{1}, \ldots, \mathbf{r}_{n}) &= \frac{1}{N} \sum\limits_{k=1}^N c_k\, \phi_k( \mathbf{r}_{1}, \ldots, \mathbf{r}_{n}) .
\end{align*}
These approximations allow us to compute kernel and path-wise generalization errors.

\subsection{Hyperparameters}
In each of the examples, the following hyperparameters are specified in the associated table: the initial conditions' probability distributions $\mathbf{\mu}_0$, the number of training trajectories $L$, the number of trajectories used to approximate the empirical data distribution $L'$, the number of time-stamps (i.e. the observation times) $J$, the training time range $T$, the generalization time range $\tilde{T}$, the uniform multiplicative noise parameter $\sigma_\text{noise}$, the number of agents $n$, the number of features $N$,  the probability distribution for the random features $\theta$, and the sparsity $s$. For the random features, the probability distribution is set to the normal distribution $\n(0,\sigma^2)$, and the tunable learning parameters are $N$, $s$ and $\sigma$ which were determined through a coarse grid search.  All codes are available at \url{https://github.com/felix-lyx/random_feature_interacting_system}.

\section{Experimental Results} \label{sec:examples}
For visualization purpose, we choose $d=2$ for our examples. In all tests, we use the random radial expansion where the trained coefficients are obtained using the least squares method (RRE) or using the sparse approximation (SRRE). The main indication of success is based on the path-wise error, which is reported for each experiment. Note that the data is acquired over a time interval $T$ which is smaller than the total time $\tilde{T}$ used in the prediction. All of the experiment include uniform multiplicative noise on the state variables with scaling parameter $\sigma_\text{noise}$. In all tests, the algorithm does not know what the true interaction kernels are and thus must learn them from observations of the dynamics. We test the two approaches on known interacting kernels so that we can provide empirical error bounds.

\begin{figure}[h!]
\centering
 \includegraphics[scale = 0.22]{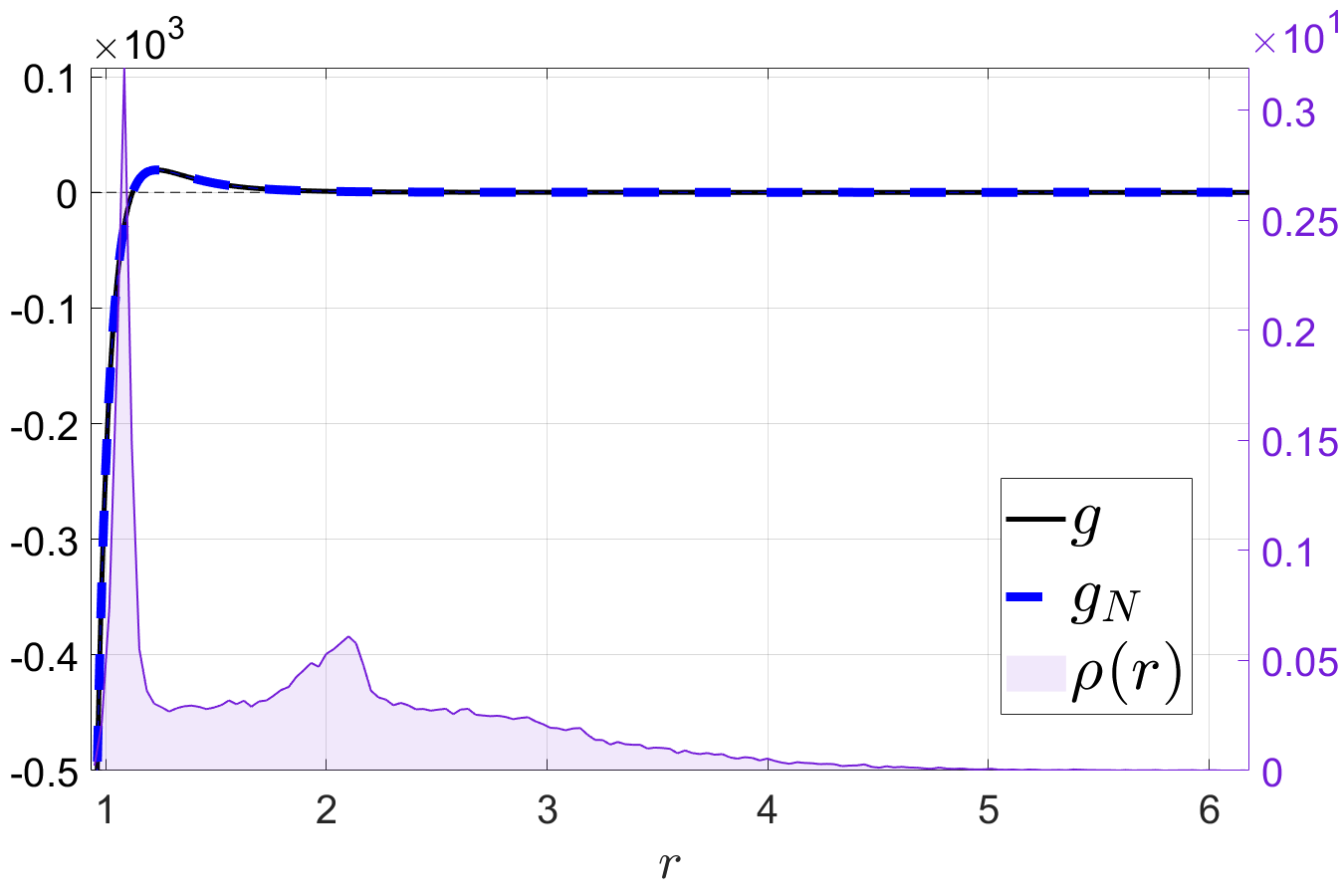}
  \includegraphics[scale = 0.22]{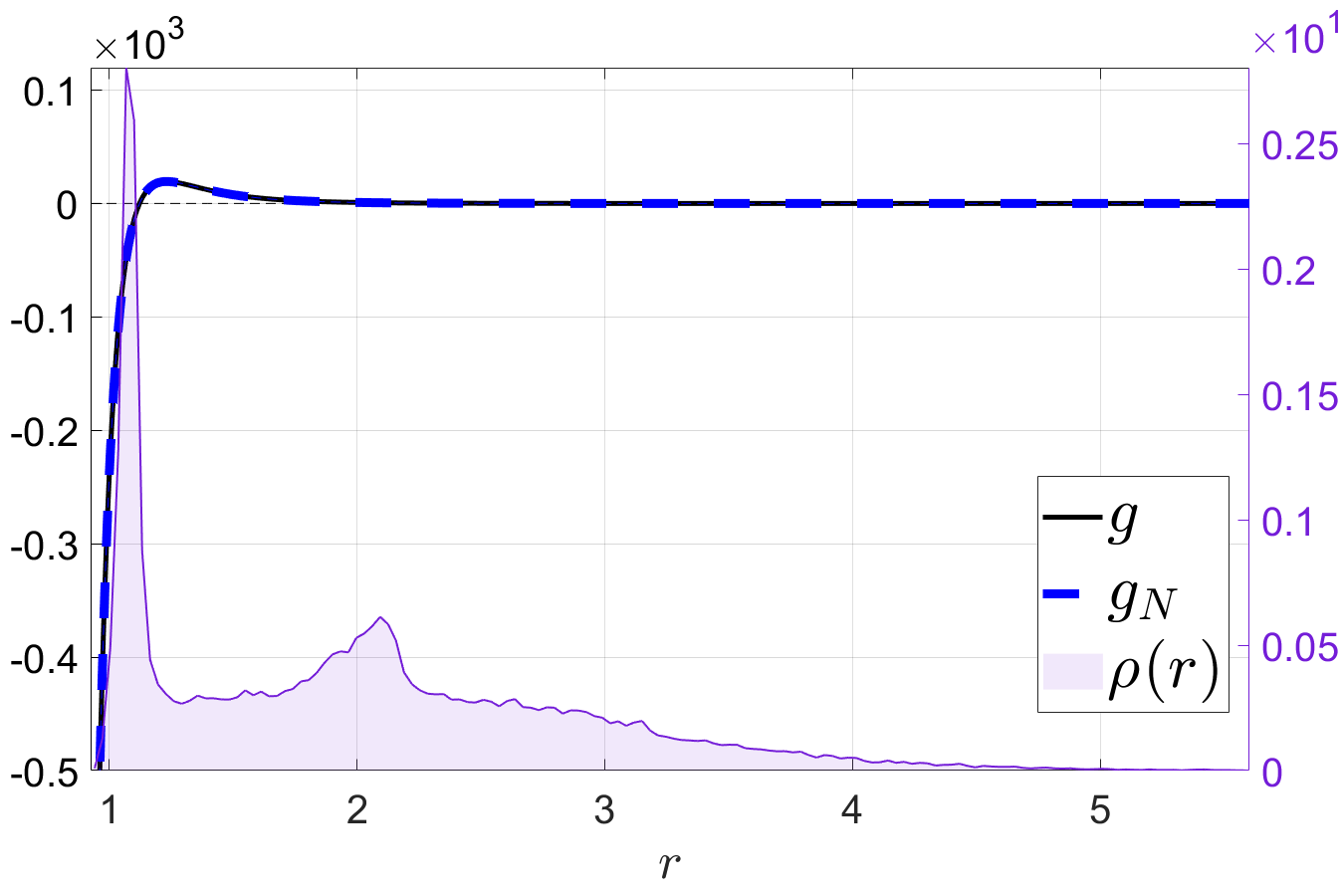}
    \caption{Lennard--Jones System: Plots of the true interaction kernel (the black curves) and the learned interaction kernels (the blue curves). The first graph uses the RRE model and the second uses the SRRE model.  }
    \label{fig:LJ_kernels}
\end{figure}

\begin{figure}[h!]
\centering
 \includegraphics[scale = 0.4]{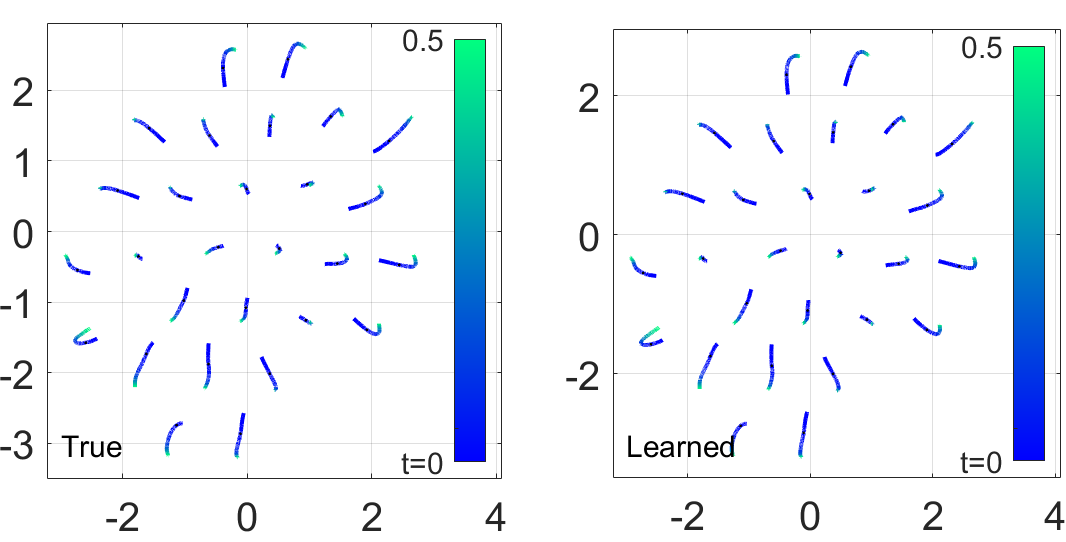}
 \includegraphics[scale = 0.4]{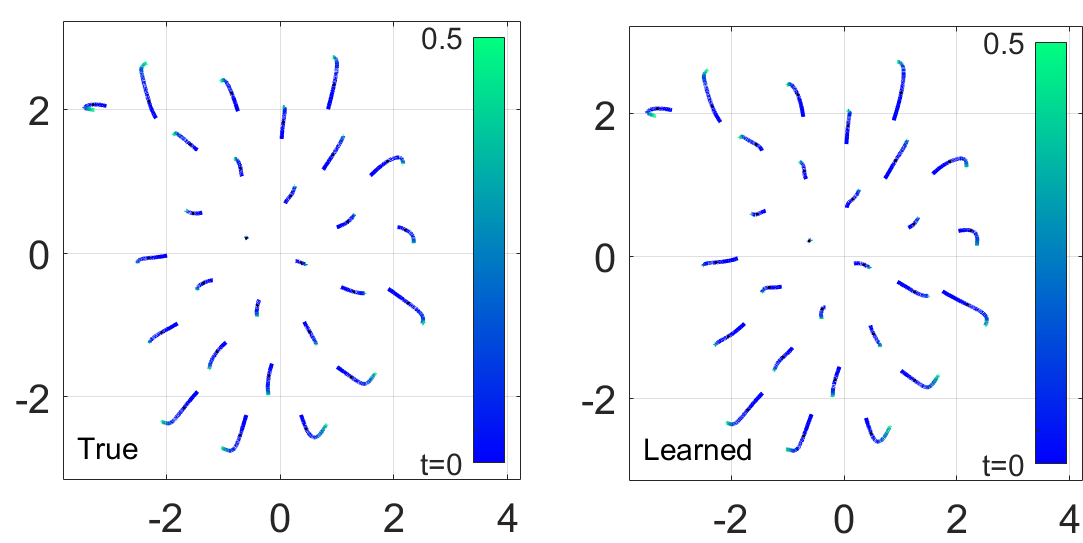}
    \caption{Lennard--Jones System: Comparison of the true paths (left column) and the learned paths (right column, top row is RRE and the bottom row is SRRE) on the test data using the trained interacting kernels. Each curve is the path of one agent in time (denoted by the colorbar).}
    \label{fig:LJ}
\end{figure}

\vspace{0.5em}
\noindent\textbf{The Lennard--Jones system} is a first-order homogeneous interacting system that models intermolecular dynamics. The governing equations are given by the following system of ODEs \cite{jones1924determination, jones1924determination2, lennard1931cohesion}: 
\begin{align}
\begin{cases}
    &\frac{d}{dt} {\mathbf x}_i(t) = \frac{1}{n} \sum\limits_{i'=1}^n g(\Vert    \mathbf{r}_{i',i}(t)\Vert_2)\, \mathbf{r}_{i',i}(t)\\
    &\mathbf{r}_{i',i}(t) =  \mathbf{x}_{i'}(t)-\mathbf{x}_i(t)
    \end{cases}
\end{align}
where $g(r)= \frac{G'(r)}{r}$ and $G(r)$ is the Lennard--Jones potential given by \begin{equation}
    G(r) = 4\ep \left[\left(\frac{\sigma}{r}\right)^{12}-\left(\frac{\sigma}{r}\right)^6\right],
\end{equation}
where $\ep=10$ is the depth of the potential well (dispersion energy) and $\sigma=1$ is the distance at which the potential between particles is 0. The $r^{-12}$ term accounts for repulsion at short ranges and the $r^{-6}$ term accounts for attraction at long ranges. {The trajectories form a specific configuration of states, which can limit the available information when training on observations of the configuration (see Figure~\ref{fig:LJ} where each curve is the path of one agent in time). This is observed numerically in Figure~\ref{fig:LJ_kernels}, where the empirical density of radial distances $\rho(r)$ (the purple region) concentrates around $r=1$ with very few measurements outside of $r\in\left[\frac{1}{2},5\right]$.} Even though the data distribution does not provide a rich sampling of $r$, the RRE (blue curve on the left) and SRRE (blue curve on the right) provide close approximations within the entire interval of interest. Both the RRE and SRRE approximations produce similar relative kernel errors of  $8.64\cdot 10^{-4}$  and $3.00\cdot 10^{-4}$, respectively. Using the learned kernels from Figure~\ref{fig:LJ_kernels}, in Figure~\ref{fig:LJ} the simulated paths are compared to the true paths starting from new initial data. The total prediction time is $50\times$ longer than the interval in which we observe the training trajectories.  The path-wise errors using random ICs are $2.22\cdot 10^{-2}\pm 6.48\cdot 10^{-2}$  (RRE)  and $8.20\cdot 10^{-3}\pm 2.44\cdot 10^{-2}$ (SRRE). The learned paths require at least one query of the learned kernels for each time-step, thus the SRRE approach leads to significant improvements in run time cost for forecasting and predictions. For this experiment, the error is slightly lower in the SRRE case with a $6.67\times$ speed up in queries at each step. Table \ref{tb:LJ_hyper} contains the hyperparameters used for this experiment.   

\begin{table}
\begin{center}
\begin{tabularx}{\columnwidth}{|c|c|C|C|c|c|c|C|}
    \hline
    $n$ & $\mathbf{\mu}_0$ & $L$ & $L'$ & $J$ & $T$ & $\tilde{T}$ & $ \sigma_\text{noise}$\\
    \hline
    7 & $\n(\mathbf{0},I_{2n})$ & 100 & 2000 & 150 & 0.01 & 0.5 & 0.001\\
    \hline
\end{tabularx}
\begin{tabularx}{\columnwidth}{|C|C|C|}
    \hline
    $N$ & $\theta$ & $s$\\
    \hline
    1000 & $\n(0,35)$ & 150\\
    \hline
\end{tabularx}
\caption{\label{tb:LJ_hyper} Hyperparameters used in the Lennard--Jones system example.}
\end{center}
\end{table}

\vspace{0.5em}

\begin{figure}[h!]
\centering
 \includegraphics[scale = 0.4]{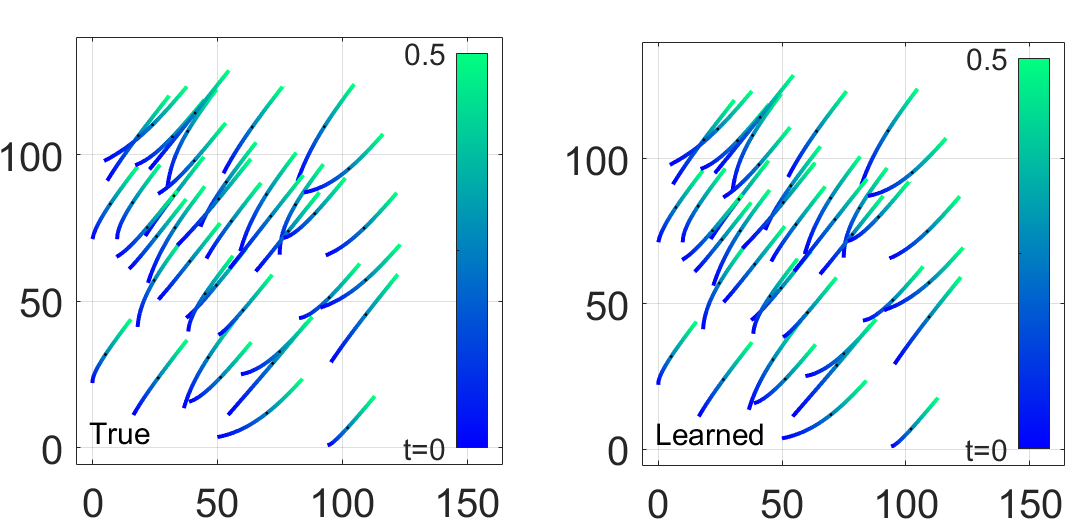}\\
 \includegraphics[scale = 0.4]{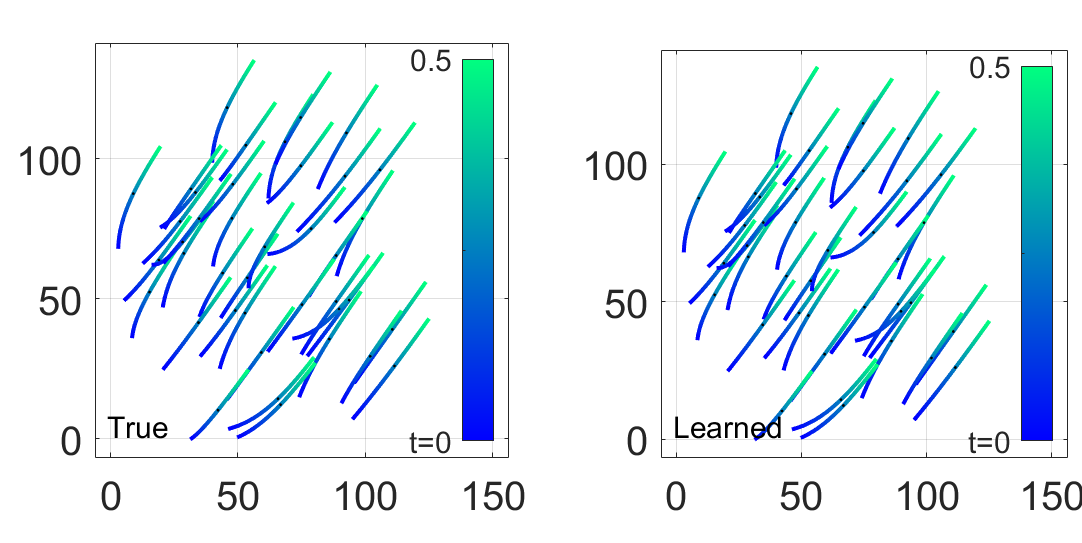}
    \caption{Cucker--Smale System: Comparison of the true paths (left column) and the learned paths (right column, top row is RRE and the bottom row is SRRE) on the test data using the trained interacting kernels. Each curve is the path of one agent in time (denoted by the colorbar).}
    \label{fig:CS}
\end{figure}

\noindent\textbf{The Cucker--Smale system} is a second-order homogeneous system. The dynamics are governed by the following systems of ODEs \cite{cucker2007emergent, cucker2007mathematics}: 
\begin{align}
\begin{cases}
    &\frac{d^2}{dt^2} {\mathbf x}_i(t) = \frac{1}{n} \sum\limits_{i'=1}^n g(\Vert    \mathbf{r}_{i',i}(t)\Vert_2)\, \frac{d}{dt}\mathbf{r}_{i',i}(t)\\
    &\mathbf{r}_{i',i}(t) =  \mathbf{x}_{i'}(t)-\mathbf{x}_i(t)
    \end{cases}
\end{align} where $g(r) = (1+r^2)^{-\frac14}$ is an alignment-based interaction kernel. Table \ref{tb:CS_hyper} contains the hyperparameters used for the system. In this setting, the RRE obtains a smaller relative kernel error than the SRRE,  that is $1.30\cdot 10^{-3}$  and $1.74\cdot 10^{-1}$, respectively, since the SRRE kernel produces a small oscillation within the support set of the distribution of observed radial distance. This is likely due to an instability formed from the second order model which can be resolved by increasing the sparsity. However, the learned and true trajectories shown in Figure~\ref{fig:CS} indicate that the overall model agrees with the true governing system. In fact, path-wise errors using random ICs are close to each other: $6.39\cdot 10^{-1}\pm 5.40\cdot 10^{-2}$  (RRE)  and $6.36\cdot 10^{-1}\pm 5.58\cdot 10^{-2}$ (SRRE). 
\begin{table}
\begin{center}
\begin{tabularx}{\linewidth}{|C|C|C|C|C|C|C|}
    \hline
    $n$  & $L$ & $L'$ & $J$ & $T$ & $\tilde{T}$ & $ \sigma_\text{noise}$\\
    \hline
    10  & 50 & 2000 & 200 & 0.25 & 0.5 & 0.01\\
    \hline
\end{tabularx}
\begin{tabularx}{\linewidth}{|c|C|C|C|}
\hline
$\mathbf{\mu}_0 = \mathbf{\mu}_{0,\text{velocity}}$ & $N$ & $\theta$ & $s$\\
\hline
$\text{Unif. on }[0,100]^2$ & 1000 & $\n(0,1)$ & 500\\
\hline
\end{tabularx}
\caption{\label{tb:CS_hyper} Hyperparameters used in the Cucker--Smale system example.}
\end{center}
\end{table}

\begin{figure}[h!]
\centering
 \includegraphics[scale = 0.22]{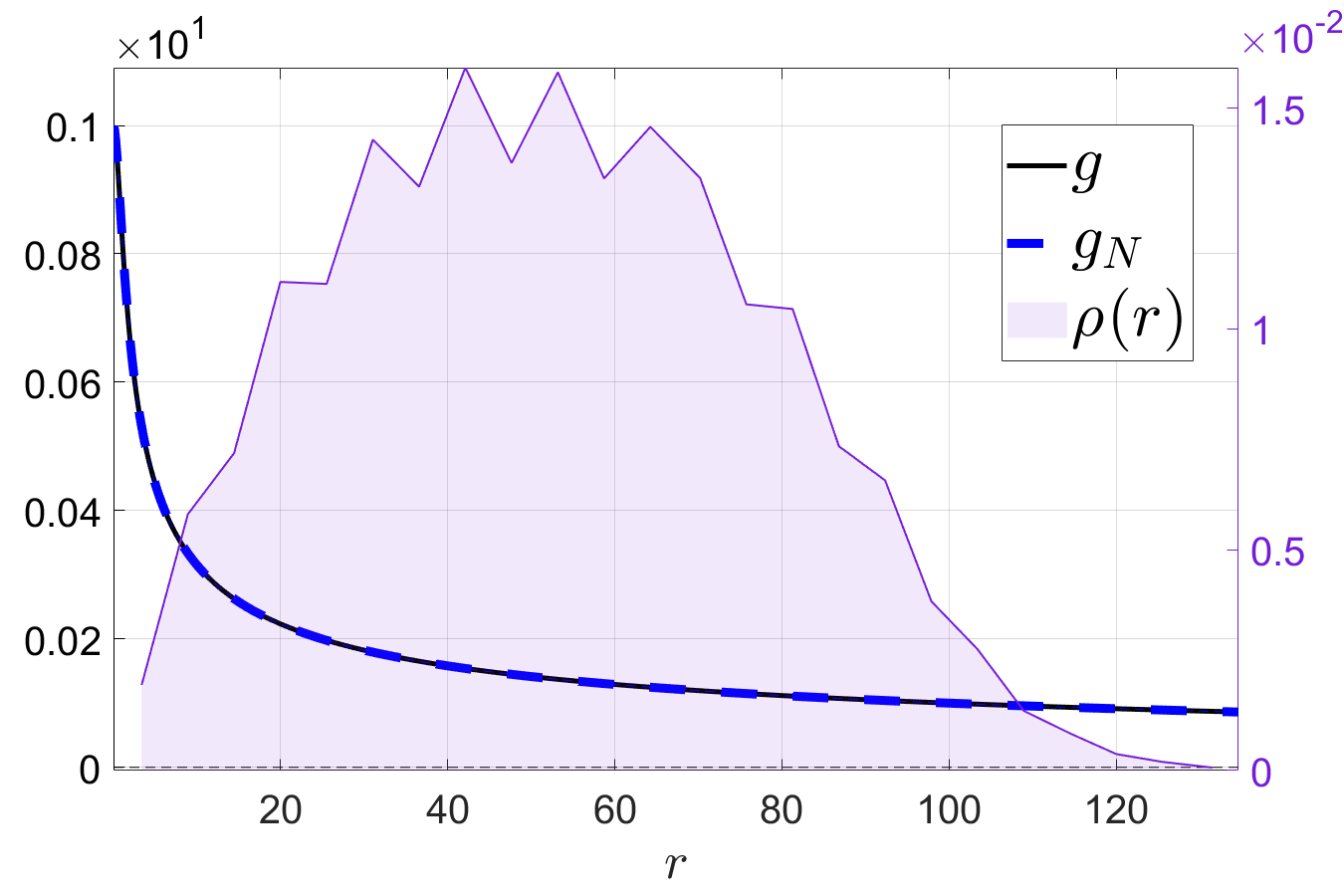}
   \includegraphics[scale = 0.22]{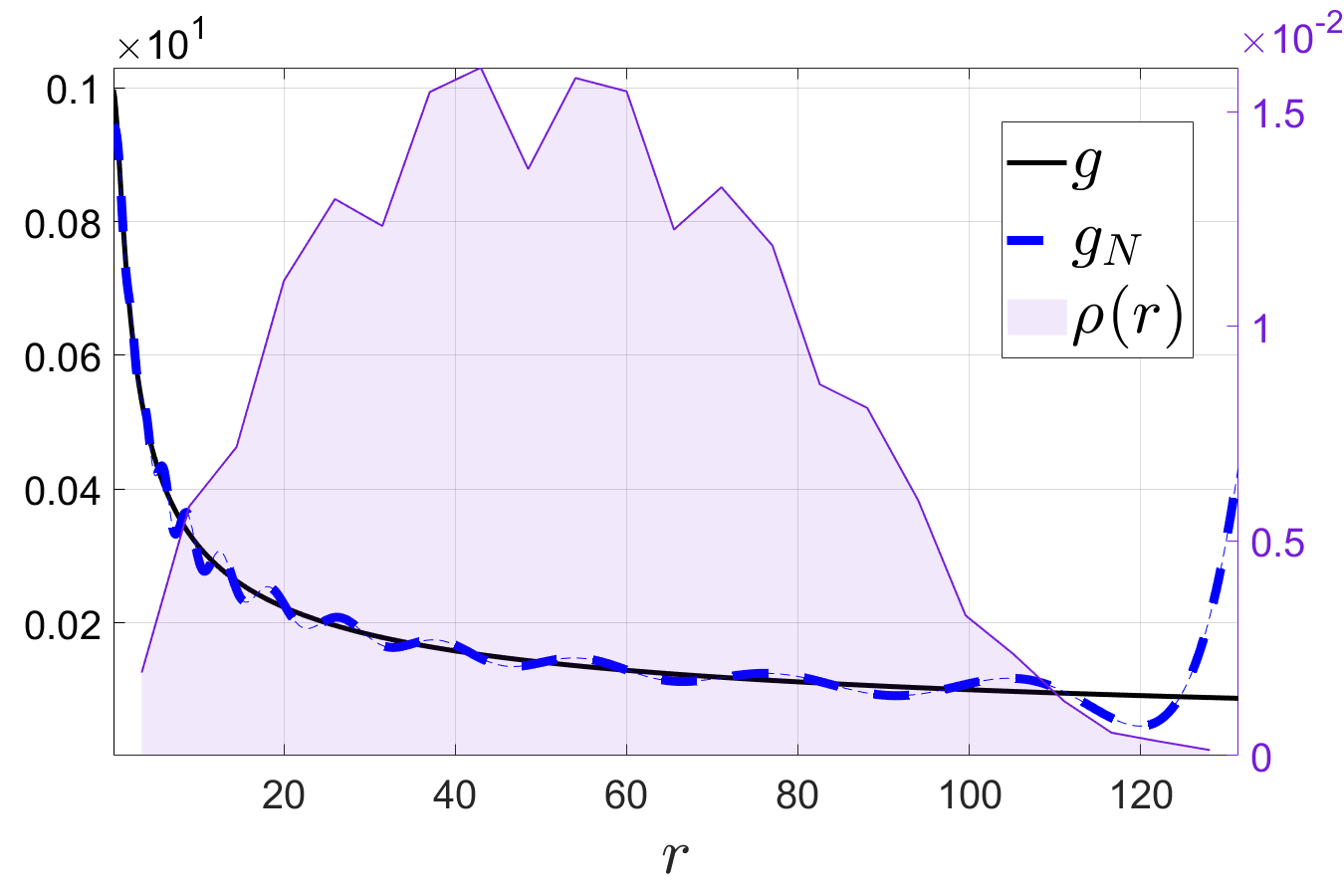}
    \caption{Cucker--Smale System: Plots of the true interaction kernel (the black curves) and the learned interaction kernels (the blue curves). The first graph uses the RRE model and the second uses the SRRE model. }
    \label{fig:CS_kernels}
\end{figure}

\vspace{0.5em}
\noindent\textbf{Single Predator-Swarming Prey Interactions} is an example of a first-order heterogeneous particle system with four interaction kernels. Each agent $\mathbf{x}_i$ has a type label $k_i\in \{1,2\}$. We use $n_1$ and $n_2$ to denote the number of agents of type $1$ (prey) and $2$ (predator), and $g_{k_ik_{i'}}$ is the interaction kernel governing how agents of type $k_{i'}$ influence agents of type $k_i$. The dynamics are governed by the following system: 
\begin{align}\label{eq:first_heterog_system}
\begin{cases}
    &\frac{d}{dt} {\mathbf x}_i(t) = \sum\limits_{i'=1}^n \frac{1}{n_{k_{i'}}} g_{k_ik_{i'}}(\Vert    \mathbf{r}_{i',i}(t)\Vert_2)\, \mathbf{r}_{i',i}(t)\\
    &\mathbf{r}_{i',i}(t) =  \mathbf{x}_{i'}(t)-\mathbf{x}_i(t)
    \end{cases}
\end{align}
where the four interaction kernels (prey-prey, prey-predator, predator-prey, predator-predator) are given by
\begin{center}
\begin{tabular}{l l}
    $g_{11}(r) = 1-r^{-2}$,  & $g_{12}(r) = -2r^{-2}$,\\ $g_{21}(r) = 3.5r^{-3}$,  & $g_{22}(r) = 0$,
\end{tabular}
\end{center}
and all terms will be learned.
One of the potential difficulties with this example is that the path-wise error depends on learning each of the kernels to within the same accuracy. In addition, the density of radial distances between the predator-prey (and by symmetry the prey-predator) concentrates at $r=1.5$ with a rapid decay outside of the peak. This example tests the methods ability to obtain accurate models with a limited sampling distribution.

\begin{figure}[h!]
\centering
 \includegraphics[scale = 0.4]{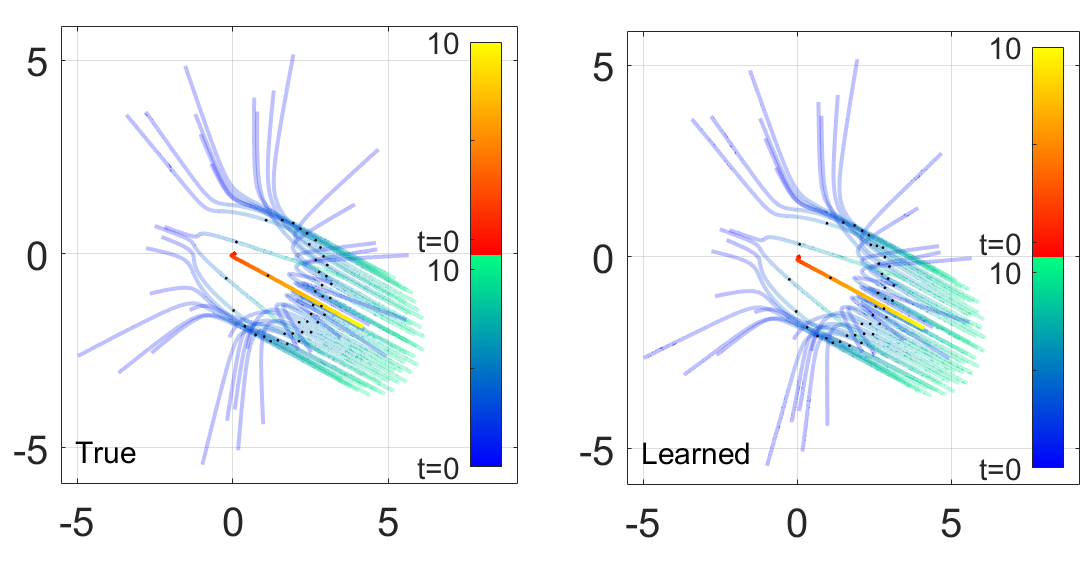}
 \includegraphics[scale = 0.4]{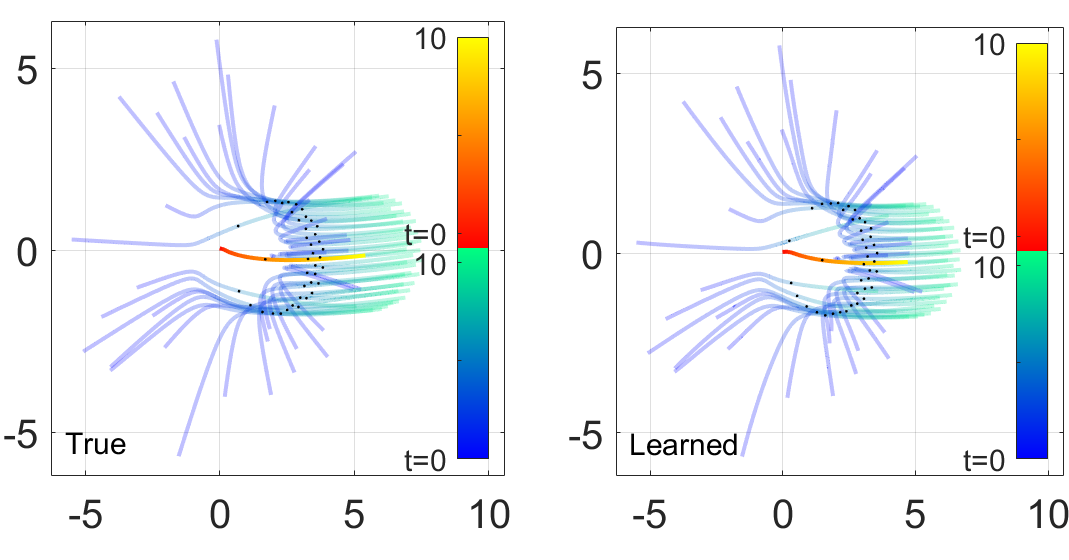}
    \caption{Single Predator-Swarming Prey Interactions:  Comparison of the true paths (left column) and the learned paths (right column, top row is RRE and the bottom row is SRRE) on the test data using the trained interacting kernels. Each curve is the path of one agent in time (denoted by the colorbar). The warm tone is the predator and the cool tone is used for the prey. The dots represent the transition point between the length of time used in the training set and the additional time in the test. }
    \label{fig:PP}
\end{figure}

Both the RRE and SRRE produce similar relative kernel errors for all kernels, with their path-wise errors using random ICs being comparable: $3.70\cdot 10^{-2}\pm 2.49\cdot 10^{-1}$ for RRE  and $8.19\cdot 10^{-3}\pm 4.63\cdot 10^{-2}$ for SRRE. The true and learned trajectories are plotted in Figure~\ref{fig:PP}, where the dots represent the transition point between the length of time used in the training set and the additional time ``extrapolated'' by the learned dynamical system. The data in Figure~\ref{fig:PP} is a new set of conditions that are not from the training set. Table \ref{tb:PS_hyper} contains the hyperparameters used for the experiment. 
\begin{table}
\begin{center}
\begin{tabularx}{\linewidth}{|C|C|C|C|C|C|C|C|}
    \hline
    $n_1$ & $n_2$ & $L$ & $L'$ & $J$ & $T$ & $\tilde{T}$ & $ \sigma_\text{noise}$ \\
    \hline
    9 & 1 & 50 & 2000 & 200 & 5 & 10 & 0.001\\
     \hline
\end{tabularx}
\begin{tabularx}{\linewidth}{|C|C|}
    \hline
     $\mathbf{\mu}_0^{1}$ & $\mathbf{\mu}_0^{2}$ \\
    \hline
     Unif. on  ring $[0.5,1.5]$ & Unif. on disk at 0.1 \\
    \hline
\end{tabularx}
\begin{tabularx}{\columnwidth}{|C|C|C|}
    \hline
    $N$ & $\theta$ & $s$\\
    \hline
    $\begin{bmatrix} 
    500 & 500\\
    500 & 50
    \end{bmatrix}$ & $\n(0,30)$ & 400\\
    \hline
\end{tabularx}

\caption{\label{tb:PS_hyper} Hyperparameters used in the Single Predator-Swarming Prey system example.}
\end{center}
\end{table}

\begin{table*}[h!]
    \centering
    \begin{tabularx}{0.9\linewidth}{|c|C|C|}
        \hline    
         & \textbf{RRE} & \textbf{SRRE}\\
        \hline
        \textbf{Relative Kernel Error} & $\begin{bmatrix}
        9.35\cdot 10^{-3} & 5.93\cdot 10^{-3}\\
        0 & 0
        \end{bmatrix}$ & $\begin{bmatrix}
        2.26\cdot 10^{-2} & 1.43\cdot 10^{-2}\\
        0 & 0
        \end{bmatrix}$ \\
        \hline
        \textbf{Path-wise Error - Training} & $6.21\cdot 10^{-1}\pm 6.89\cdot 10^{-1}$ & $1.69\pm 9.23\cdot 10^{-1}$\\
        \hline
        \textbf{Path-wise Error - Testing} & $8.32\cdot 10^{-1}\pm 9.53\cdot 10^{-1}$ & $1.72\pm 9.49\cdot 10^{-1}$\\
        \hline
    \end{tabularx}
    \caption{Kernel and generalization errors of Sheep-Food interacting system example.}
    \label{table:Sheep_Errors}
\end{table*}

\begin{figure}[h!]
\centering
 \includegraphics[scale = 0.4]{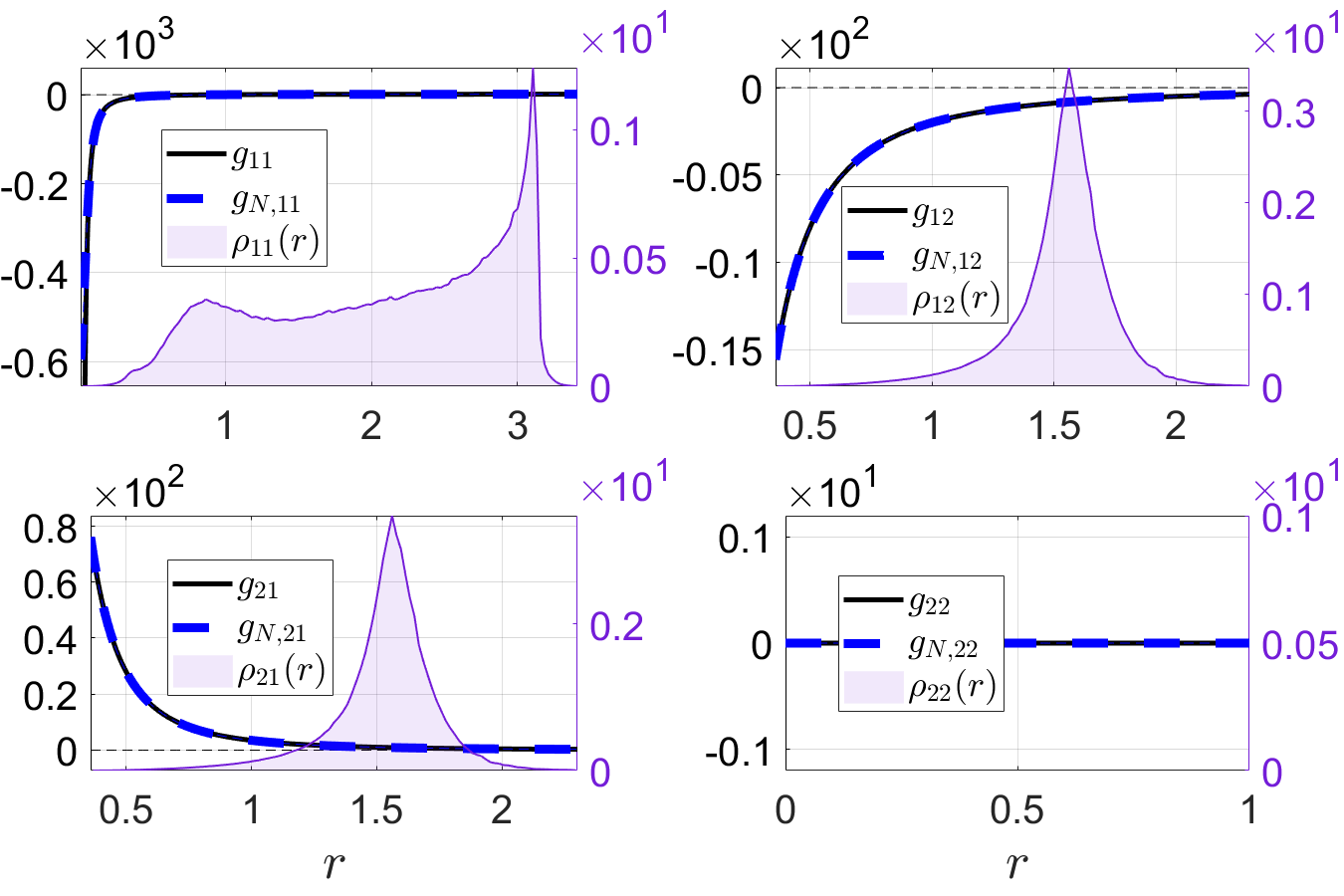}
  \includegraphics[scale = 0.4]{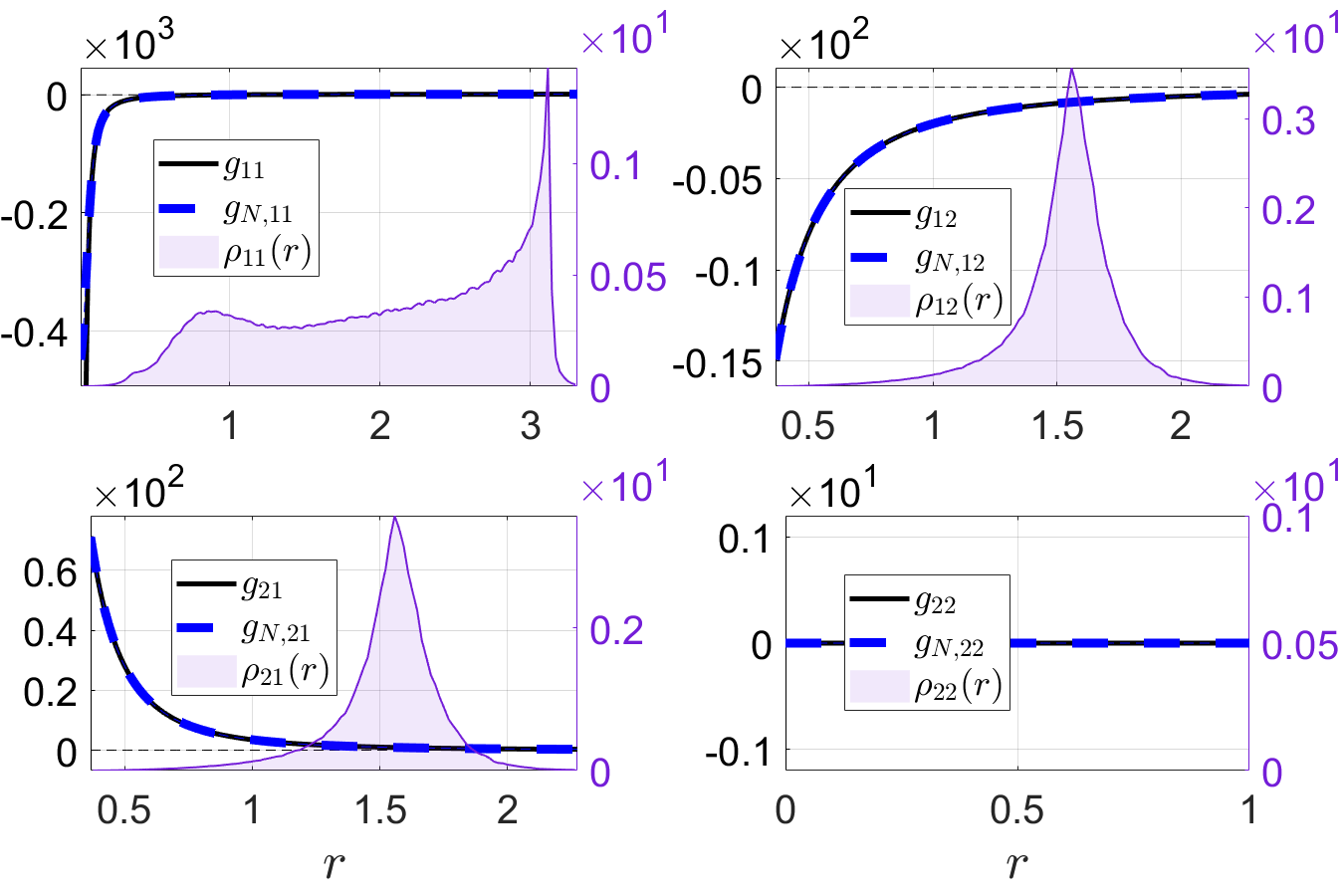}
    \caption{Single Predator-Swarming Prey Interactions: Plots of the true interaction kernels (the black curves) and the learned interaction kernels (the blue curves). The first four graphs use the RRE model and the last four graphs use the SRRE model. }
    \label{fig:PP_kernels}
\end{figure}

\begin{figure}[h!]
\centering
 \includegraphics[scale = 0.4]{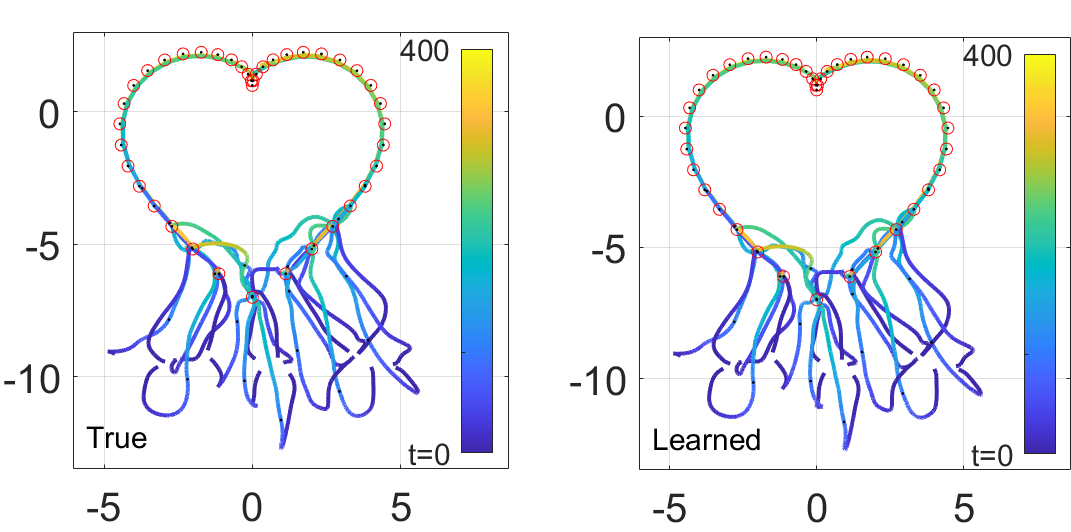}
 \includegraphics[scale = 0.4]{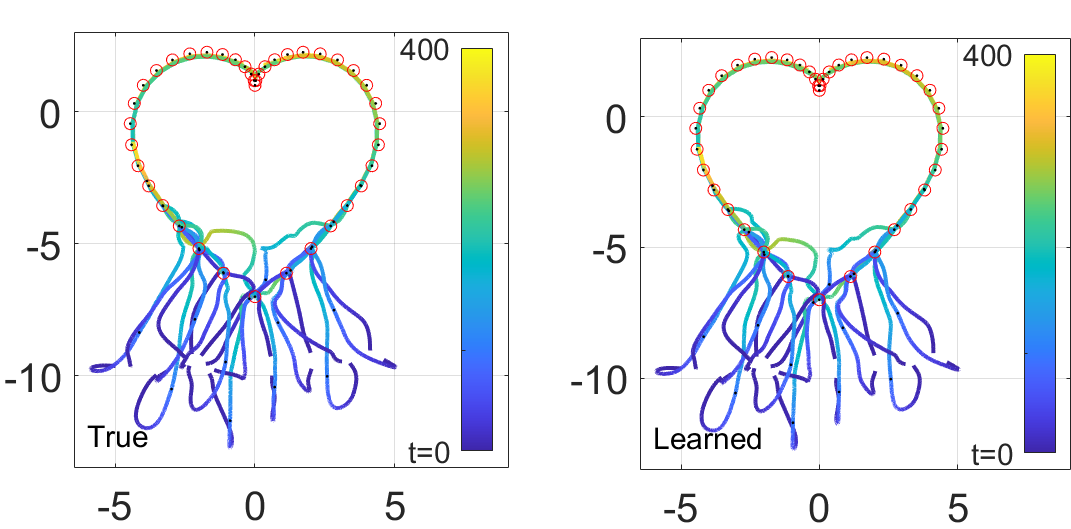}
    \caption{Sheep-Food Interacting System: Comparison of the true paths (left column) and the learned paths (right column, top row is RRE and the bottom row is SRRE) on the test data using the trained interacting kernels.}
    \label{fig:Sheep_Kernel}
\end{figure}

\begin{figure}[h!]
\centering
 \includegraphics[scale = 0.4]{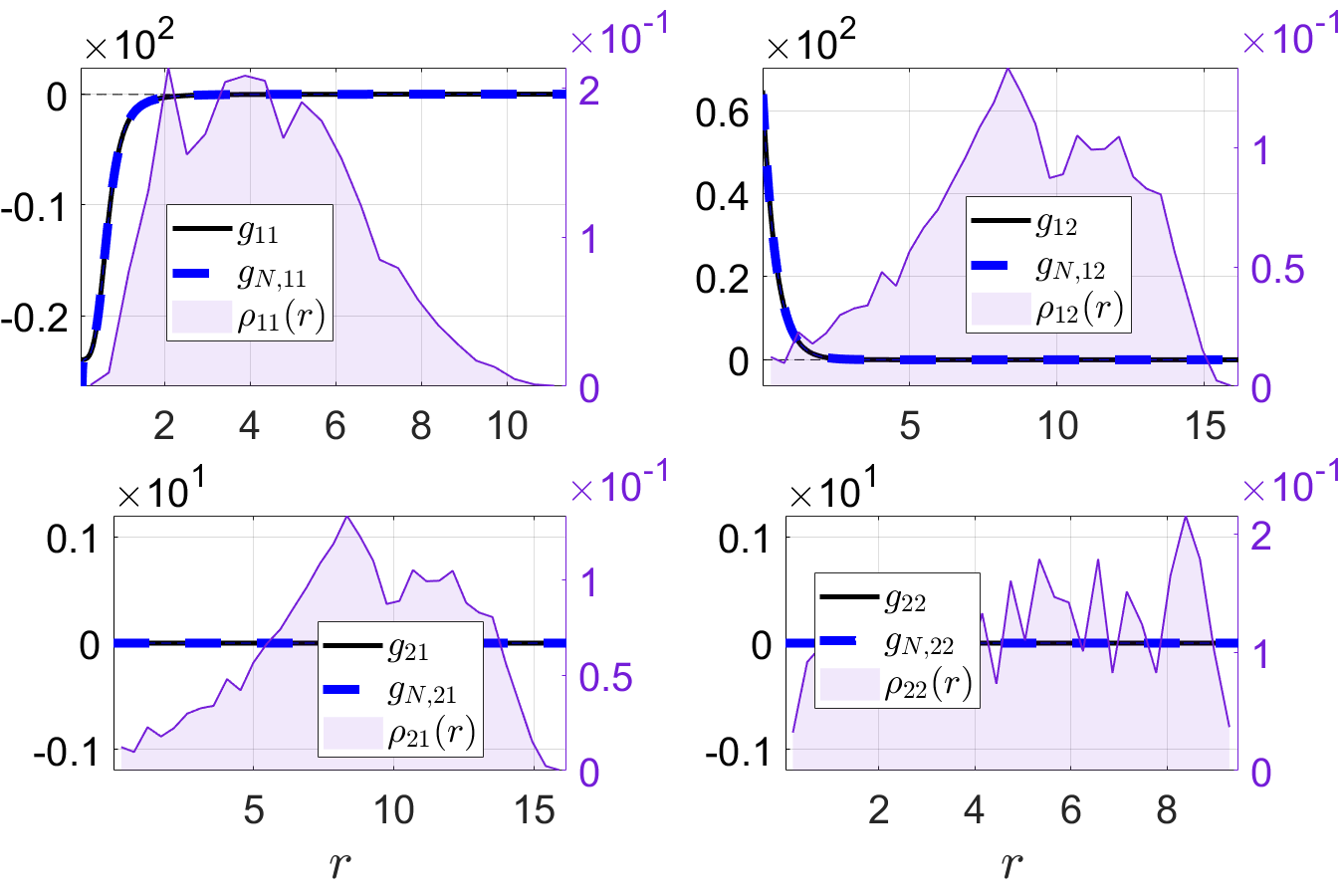}
   \includegraphics[scale = 0.4]{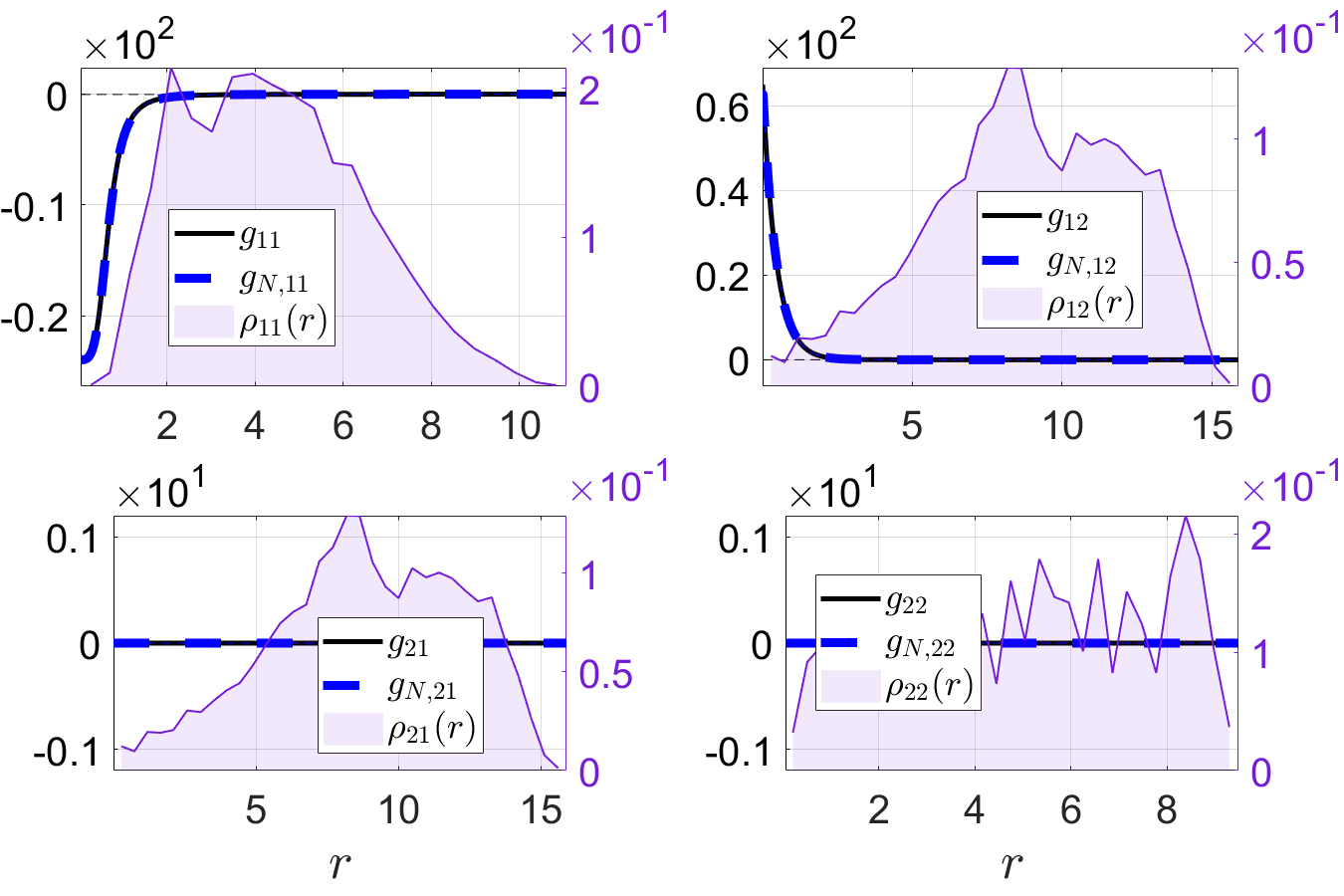}
    \caption{Sheep-Food System: Plots of the true interaction kernels (the black curves) and the learned interaction kernels (the blue curves). The first four graphs use the RRE model and the last four graphs use the SRRE model. }
    \label{fig:Sheep}
\end{figure}

\vspace{0.5em}
\noindent\textbf{Sheep-Food Interacting System:} Inspired by art which captures large flocks of sheep swarming various grain patterns  \cite{Harkins_WP}, we propose a sheep-food interacting model and try to discover the interactions based on observations.  Our approach to model the sheep is based off of the prey dynamics in \cite{chen2014minimal}. Similarly, the sheep take on the role of predator in their interaction with the stationary food source.  The sheep both want to cluster together for safety, but also are drawn to seek out the food at the cost of the group cohesion. We manually set the  dynamics to be governed by a first-order heterogeneous system:  $\eqref{eq:first_heterog_system}$ where \begin{center}
\begin{tabular}{l l}
    $g_{11}(r) = -\frac{1.2}{r^4/4+0.05}+0.02$,  & $g_{12}(r) = 65e^{-\frac{r}{0.45}}$,\\ $g_{21}(r) = 0$,  & $g_{22}(r) = 0$.
\end{tabular}
\end{center}
{where type $1$ is the food and type $2$ are the sheep.} The food is the stationary ``prey'' and are arranged in a ``heart'' shape and are plotted as targets in Figure~\ref{fig:Sheep}. The sheep (the predator swarm) are initialized randomly below the food, near $y=-10$. The challenge is to correctly learn the dynamics from the earlier time interactions in order to correctly forecast the entire path the sheep take around the heart configuration. Note that although the training set only includes the interaction in the blue region, up to $T=100$, the forecasted paths capture the full geometry. The errors for the kernels and paths are displayed in Table~\ref{table:Sheep_Errors}. The RRE approximation produces smaller errors but the overall kernels are similar, see Figure~\ref{fig:Sheep_Kernel}.

\begin{table}[t!]
\centering
\begin{tabularx}{\linewidth}{|C|C|C|C|C|C|c|C|}
    \hline
    $n_1$ & $n_2$ & $L$ & $L'$ & $J$ & $T$ & $\tilde{T}$ & $ \sigma_\text{noise}$ \\
    \hline
    20 & 40 & 50 & 1000 & 600 & 100 & 400 & 0.001\\
     \hline
\end{tabularx}
\begin{tabularx}{\linewidth}{|c|C|}
    \hline
     $\mathbf{\mu}_0^{1}$ & $\mathbf{\mu}_0^{2}$ \\
    \hline
     Unif. on $[-9,-10]{\times} [-5,5]$ & Heart of ``radius'' 5 at $\mathbf{0}$ \\
    \hline
\end{tabularx}
\begin{tabularx}{\columnwidth}{|C|C|C|}
    \hline
    $N$ & $\theta$ & $s$\\
    \hline
    $\begin{bmatrix}
    500 & 500\\
    50 & 50
    \end{bmatrix}$ & $\n(0,10)$ & 600\\
    \hline
\end{tabularx}
\caption{\label{tb:SF_hyper} Hyperparameters used in the sheep-food interacting system example.}
\end{table}

\vspace{0.5em}
\noindent\textbf{Comparison:} Figure~\ref{fig:compare_with_Fourier} provides a comparison of our approach and the standard RFF \cite{rahimi2007random, rahimi2008uniform, rahimi2008weighted}, showing that our randomized feature construction produces a more accurate approximation to the interacting kernels.

We compare our algorithm on the Lennard--Jones example with the approximation produced by the piecewise polynomial model from \cite{lu2019nonparametric}. The setup is as described in the previous Lennard--Jones example. The model from \cite{lu2019nonparametric} using $L=1000$ initial conditions for the training set produces a relative kernel error of $2.26\cdot 10^{-2}$ and a path-wise generalization error of $7.31\cdot 10^{-2}$. The SRRE using $L=100$ initial conditions for the training set produces a relative kernel error of $3.00\cdot 10^{-4}$ and a path-wise generalization error of $8.20 \cdot 10^{-3}$. This highlights two advantages of our approach. The RRE and SRRE methods need an order of magnitude fewer samples since they use a global smooth candidate set rather than a local basis. Secondly, the generalization error is lower due to the regularizing nature of the method, i.e. only retaining the dominate random features.

\section{Discussion}

We developed a random feature model and learning algorithm for approximating interacting multi-agent systems. The method is constructed using an integral based model of the interacting velocity field which is approximated by a RFM directly on observations of state variables. The resulting functions, i.e. the sums-of-Gaussians, led to a new radial randomized feature space which is theoretically justified and was shown to produce a more consistent and stable result than other RFMs. We employ a randomized feature space rather than a pre-defined set of candidate functions, so that the algorithm would not require prior knowledge on the system or its interaction kernels. This is motivated by the applications, since manually modeling and training complex multi-agent systems is often computationally infeasible.

We used a sparse random feature training algorithm based on the HTP algorithm. We note that when the sparsity parameter is set to the feature space size, i.e. $s=N$, the training reverts to the least squares problem and is applicable for some of the problems examined in this work. In particular, when one has a sufficiently rich dataset, least squares based RFMs produce accurate solutions.  On the other hand, it is likely that there is only a limited set of observations of the system. In that case there are several benefits to incorporating sparsity in the training of RFM, i.e.  extracting a RFM that only uses a small number of randomized features. First, sparsity lowers the cost of forecasting for high-dimensional multi-agent system, since the cost scales with $s$ rather than $N$. Since the trained systems are used to forecast the multi-agent dynamical system, a large number of repeated queries are needed, which would limit the use of standard RFM and overparameterized neural networks. Sparse regression also leads to less overfitting when one has a scarce or limited training set. Lastly, although SRRE may lead to models which are less interpretable than fixed candidate dictionaries such as \cite{brunton2016discovering,schaeffer2017learning,peherstorfer2016data}, the SRRE approach does provide some insight into the structure of the interaction kernel and can be used to guide analysis of the underlying system.

Our approach obtained accurate predictions for popular first-order system with homogeneous and heterogeneous interactions, second-order systems, and a new sheep-food interacting system from noisy measurements. A comparison with a similar approach showed that the SRRE produced lower path-wise generalization error and lower kernel error while also needing an order of magnitude fewer training points. As with most approximation techniques for training dynamical systems, one limitation of the SRRE is with high levels of noise and outliers. One way to alleviate this is to preprocess the data, possibly using a neural network model, to approximate the underlying training paths before extracting its collective behavior. However, the choice of preprocessing algorithms may bias the trained dynamics. Another limitation is its inability to train highly accurate predictive models when the underlying interacting system incorporates additional terms outside of the assumptions. A future direction may be to train only the interacting kernel component of the system and develop a closure approach for obtaining the remaining aspects of the dynamics.

\section*{Acknowledgements}

S.G.M. would like to acknowledge the support of NSF grants $\#1813654$, $\#2112085$  and the Army Research Office (W911NF-19-1-0288). H.S. was supported in part by AFOSR MURI FA9550-21-1-0084 and NSF DMS-1752116.

\clearpage
\newpage
\bibliographystyle{plain}

\end{document}